%% file: neurips_2025.tex
\documentclass{article}

    \usepackage[final]{neurips_2025}

\usepackage[colorlinks=true, linkcolor=red, urlcolor=black, citecolor=blue, anchorcolor=blue, backref=page]{hyperref}
\renewcommand*{\backref}[1]{}
\renewcommand*{\backrefalt}[4]{%
    \ifcase #1%
          \or [Cited on page~#2.]%
          \else [Cited on pages~#2.]%
    \fi%
    }
\usepackage{comment}
\usepackage{natbib}
\usepackage{times}
\usepackage{latexsym}
\usepackage{bm}

\usepackage[utf8]{inputenc} 
\usepackage[T1]{fontenc}    
\usepackage{hyperref}       
\usepackage{url}            
\usepackage{booktabs}       
\usepackage{amsfonts}       
\usepackage{nicefrac}       
\usepackage{microtype}      
\usepackage{xcolor}         
\usepackage{xspace}
\usepackage{pifont}
\usepackage{tikz}
\usetikzlibrary{bayesnet}
\usepackage{wrapfig}
\usepackage{lipsum}

\hypersetup{
    colorlinks=true,
    linkcolor=blue,
    }

\usepackage{inconsolata}
\usepackage{amsmath}
\usepackage{mathtools}
\usepackage{graphicx}
\usepackage{centernot}
\usepackage{amsfonts}
\usepackage{booktabs}

\usepackage{enumitem}
\usepackage{soul}
\usepackage{lineno}
\usepackage{algorithm}
\usepackage{algorithmic}
\usepackage{multirow}
\usepackage{wrapfig}
\usepackage{rotating,lipsum,graphicx}

\usepackage{titletoc}

\newcommand\openweightmodelcount{$28$\ }
\usepackage{cleveref}
\usepackage{bbm}
\crefname{figure}{Figure}{Figures}
\crefname{table}{Table}{Tables}
\crefname{appendix}{Appendix}{Appendices}
\crefname{section}{Section}{Sections}
\crefname{equation}{Eq.}{Eqs.}
\crefname{enumi}{}{} %

\usepackage[utf8]{inputenc} 
\usepackage[T1]{fontenc}    
\usepackage{hyperref}       
\usepackage{url}            
\usepackage{booktabs}       
\usepackage{amsfonts}       
\usepackage{nicefrac}       
\usepackage{microtype}      
\usepackage{xcolor}         

\usepackage{titletoc}
\usepackage{multicol}
\usepackage{multirow}

\usepackage{amssymb}
\usepackage{pgfplots}
\usepackage{etoc} 

\usepackage[normalem]{ulem}  

\usepgfplotslibrary{patchplots}
\usetikzlibrary{patterns, positioning, arrows}
\pgfplotsset{compat=1.15}

\pgfmathsetmacro\sprayRadius{.5pt}
\pgfmathsetmacro\sprayPeriod{0.7cm}

\pgfdeclarepatternformonly{spray}{\pgfpoint{-\sprayRadius}{-\sprayRadius}}{\pgfpoint{1cm + \sprayRadius}{1cm + \sprayRadius}}{\pgfpoint{\sprayPeriod}{\sprayPeriod}}{
    \foreach \x/\y in {2/53,6/52,11/48,23/49,20/47,32/46,41/47,47/51,56/52,46/44,4/43,16/42,33/41,41/37,49/35,55/31,37/35,44/30,28/37,24/36,17/37,7/38,0/31,8/29,18/31,28/30,37/28,30/27,46/24,51/21,24/23,12/24,4/21,18/19,12/16,31/21,38/18,26/16,46/16,56/12,52/10,45/8,51/4,37/12,35/7,24/9,14/9,2/12,8/6,15/4,27/0,34/1,40/1} {
    \pgfpathcircle{\pgfpoint{(\x + random()) / 57 * \sprayPeriod}{\sprayPeriod - (\y + random()) / 55 * \sprayPeriod}}{\sprayRadius}
    }
\pgfusepath{fill}
}

\title{Geometry of Decision Making in Language Models}

\author{
{\bf Abhinav Joshi}$^\diamond$ 
\qquad {\bf Divyanshu Bhatt}$^{\mathparagraph\dagger}$\thanks{Work primarily done at IIT Hyderabad}
\qquad {\bf Ashutosh Modi}$^\diamond$ 
 \\ 
 $^\diamond$Indian Institute of Technology Kanpur  (IIT Kanpur) \\
  $^\dagger$Indian Institute of Technology Hyderabad (IIT Hyderabad)\\
        $^\mathparagraph$Samsung Research \& Development Institute, Bangalore \\
  \texttt{divyanshu.bh@samsung.com}, \\
  \texttt{\{ajoshi,ashutoshm\}@cse.iitk.ac.in}  
}

\begin{document}

\maketitle

\input{./sections/abstract}
\input{./sections/introduction}
\input{./sections/relatedworks}
\input{./sections/background}
\input{./sections/experiments}
\input{./sections/results}
\input{./sections/limitations}
\input{./sections/conclusion}

\bibliography{references}
\bibliographystyle{plainnat}
\newpage
\clearpage

\appendix
\input{./sections/appendix}




\end{document}

%% file: sections/abstract.tex
\begin{abstract}
    Large Language Models (LLMs) show strong generalization across diverse tasks, yet the internal decision-making processes behind their predictions remain opaque. In this work, we study the geometry of hidden representations in LLMs through the lens of \textit{intrinsic dimension} (ID), focusing specifically on decision-making dynamics in a multiple-choice question answering (MCQA) setting. We perform a large-scale study, with 28 open-weight transformer models and estimate ID across layers using multiple estimators, while also quantifying per-layer performance on MCQA tasks. Our findings reveal a consistent ID pattern across models: early layers operate on low-dimensional manifolds, middle layers expand this space, and later layers compress it again, converging to decision-relevant representations. Together, these results suggest LLMs implicitly learn to project linguistic inputs onto structured, low-dimensional manifolds aligned with task-specific decisions, providing new geometric insights into how generalization and reasoning emerge in language models. 
\end{abstract}

%% file: sections/introduction.tex
\section{Introduction} \label{sec:introduction}


Large Language Models (LLMs) have exhibited impressive generalization across diverse natural language tasks \citep{Radford2019LanguageMA, incontextfewshotlearners}. Despite their success, how these models internally arrive at decisions, particularly in tasks requiring structured reasoning, remains underexplored. Understanding this process is central to interpretability and may yield insights into model generalization, failure modes, and capabilities. Recent work in mechanistic interpretability has highlighted specific circuits or components underlie LLM reasoning \citep{elhage2021mathematical, olsson2022incontextlearninginductionheads}. In parallel, probing-based approaches have tracked how task-relevant information flows across layers \citep{tenney-etal-2019-bert, hewitt-manning-2019-structural}. However, these techniques often focus on how the information is represented and where it resides, rather than how the representation geometry evolves to support decision-making. To complement these perspectives, we study decision-making by analyzing geometric properties of underlying manifolds. We specifically make use of the \textit{Intrinsic Dimension} (ID), which quantifies the minimal degrees of freedom required to describe a distribution in high-dimensional space \citep{prml-bishop}. Prior work has demonstrated that neural representations often lie on low-dimensional manifolds \citep{gong2019intrinsic,valeriani2023the}, with ID fluctuations signalling transitions in learning and abstraction \citep{cheng2025emergence}. Yet, the connection between these geometric changes and model decisiveness, i.e., the commitment to a specific prediction, has not been explored extensively.

Our primary focus is to understand how internal decision-making unfolds within transformer-based LLMs, particularly in tasks requiring symbolic reasoning and choice commitment. To this end, we are guided by three key questions: \textbf{1) }\textbf{How} does ID evolve across layers, and how does this reflect the model’s progression from contextual encoding to decision-making? \textbf{2) }\textbf{Can} geometric markers, such as ID peaks, serve as interpretable indicators of decisiveness and confidence in model predictions?
\textbf{3) } \textbf{Are} these ID dynamics consistent across different model families and tasks, and what role does model size, training stage, or prompt conditioning (e.g., few-shot examples) play in shaping these trajectories?
We aim to bridge representational geometry with functional behavior in LLMs through these questions, providing a complementary perspective to circuit-based or probing-based analyses. Our findings reveal that ID can act as a proxy for representational focus and task commitment, helping identify critical layers that solidify/freeze model decisions, and provide insights that may guide future interpretability and intervention strategies.

\begin{figure*}[t]
\begin{center}
\centerline{\includegraphics[scale=0.45]{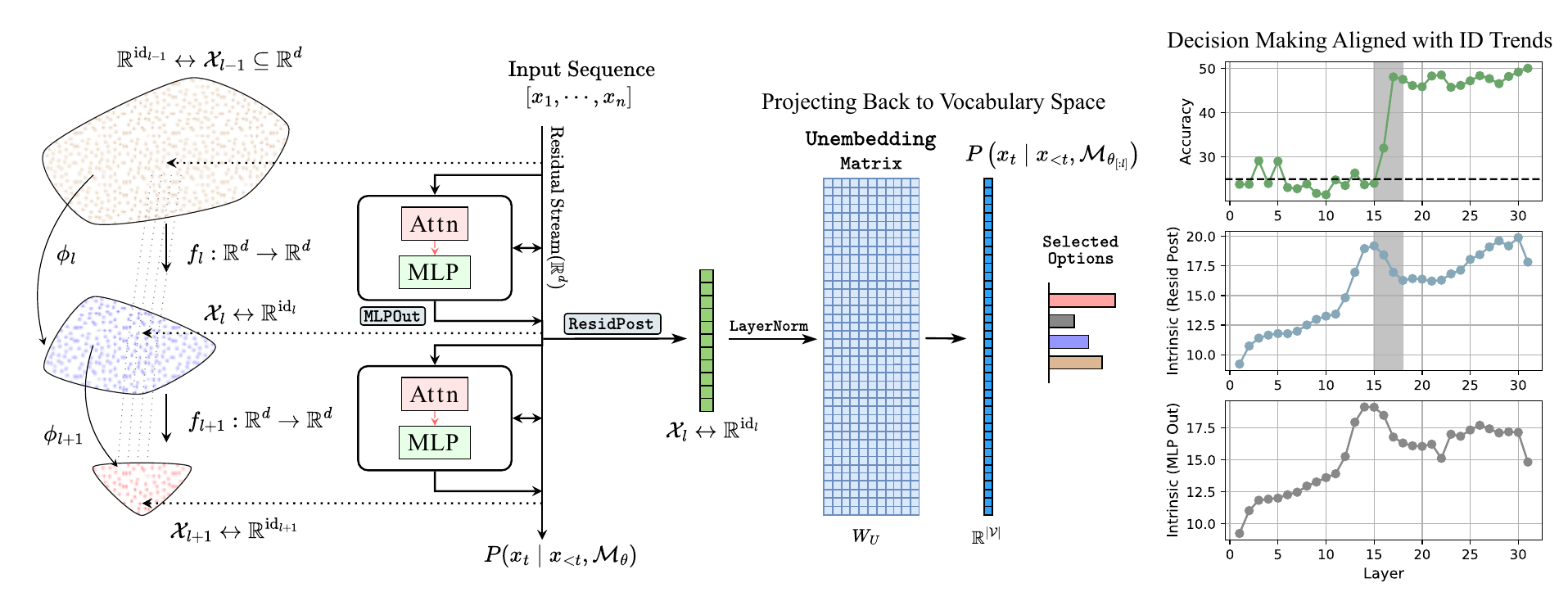}}
\caption{
In the transformer-based architectures, a vector (latent features) of the same hidden dimensions $d$, is transformed by transformer blocks $f_l$. Though the extrinsic dimension remains the same, we find that the feature space lies on low-dimensional manifolds of different intrinsic dimensions $\mathbb{R}^{\mathrm{id}_{l}}$. Intrinsically, there exists a mapping $\phi_l$ corresponding to each $f_l$, from $\mathbb{R}^{\mathrm{id}_{l-1}}\to \mathbb{R}^{\mathrm{id}_{l}}$. We study how these compressed manifolds align with the decision-making process in middle layers. We project the internal representations back to the vocabulary space to inspect the decisiveness. There is a sudden shift in performance that is aligned with the follow-up of a sharp peak observed in the residual-post ID estimates.
}
\label{fig:intrinsic-diagram-thumbnail}
\end{center}
\end{figure*}

In this work, we study the evolution of hidden representations that develop during decision making in LLMs using ID estimates by experimenting with reasoning-based multiple-choice question answering (MCQA)-style prompts. We conduct an extensive investigation into the internal representations of LLMs, analyzing \openweightmodelcount open-weight transformer models spanning multiple architectures and sizes (list of models in App. \ref{app:open-weight-models}). We build upon classical estimators such as Maximum Likelihood Estimation (MLE) \citep{MLE_id} and Two Nearest Neighbors (TwoNN) \citep{TwoNN_id}, and incorporate the recently proposed Generalized Ratios Intrinsic Dimension Estimator (GRIDE) \citep{GRIDE_id} which demonstrates improved robustness to sampling noise and curvature distortions (see \S\ref{sec:background}). 
Fig. \ref{fig:intrinsic-diagram-thumbnail} outlines our approach (details in \S\ref{sec:experiments}). 
Our primary findings are as follows: 
\begin{itemize}[nosep,noitemsep,leftmargin=*]
    \item \textbf{Emergence of Decision Geometry:} Across models and tasks, we observe a characteristic hump-shaped trend in intrinsic dimension estimates (notably at the MLP output layers), where ID increases, peaks, and then declines. This reflects an early phase of abstraction followed by convergence toward decision-specific subspaces.

    \item \textbf{ID Peaks Coincide with Decisiveness:} For most models, the peak in intrinsic dimension aligns closely with the onset of confident predictions (as revealed via projection to vocabulary space). This suggests a geometric marker of decisiveness within the model's forward pass.

    \item \textbf{Layer-Specific Dynamics Differ by Component:} We distinguish between MLP output and residual post-activations. While MLP outputs exhibit clear ID peaks and sharper reductions in later layers, residual post activations show more gradual trends, providing complementary views on when and how information/decisions solidify.

    \item \textbf{Few-Shot Prompting Sharpens Representations:} Increasing few-shot examples leads to steeper ID transitions, implying more efficient compression and faster convergence to decision-ready states.

    \item \textbf{Model Scale and Architecture Matter:} Larger models tend to reach ID peaks earlier in the layer stack and maintain lower terminal ID, hinting at more efficient abstraction and early decisiveness. Notably, model families like LLaMA and Pythia show distinct ID trends, underscoring architectural influence on representational geometry. 

\end{itemize}


In a nutshell, our study covers both real-world benchmarks and template-based tasks, enabling us to characterize representational dynamics across a diverse range of reasoning and language understanding skills. 
{We release the codebase and results at \url{https://github.com/Exploration-Lab/dim-discovery-archive}}.

%% file: sections/relatedworks.tex
\section{Related Works}

The manifold hypothesis posits that high-dimensional data often lie on low-dimensional manifolds \citep{Ruderman1994TheSO, charting-a-manifold, fefferman2013testingmanifoldhypothesis, Goodfellow-et-al-2016}. In this context, the intrinsic dimension (ID) 
has emerged as a useful geometric lens for studying neural representations. Prior work has shown that deep networks learn low-ID features \citep{gong2019intrinsic, ansuini2019intrinsic, aghajanyan2020intrinsic, pope2021intrinsic}, with lower ID correlating with better generalization \citep{gong2019intrinsic, nakada2020adaptive, aghajanyan2020intrinsic}. These trends have been well-explored in computer vision, where ID is linked to optimization geometry \citep{DBLP:journals/corr/abs-1804-08838, DBLP:journals/corr/abs-1801-02613, 8578388, zhang-generalization} and dataset complexity \citep{pope2021intrinsic, mnist, krizhevsky2009cifar, imagenet, lin2015microsoftcococommonobjects, celeba}. In transformers, early studies demonstrated similar ID patterns across layers in models trained on non-text domains like proteins and images \citep{valeriani2023the}. More recent work has brought these geometric insights into NLP. \citet{cheng2025emergence} identifies a high-ID abstraction phase in transformers, predictive of generalization and linguistic transfer. \citet{antonello2024evidence} provides complementary fMRI evidence for a two-phase abstraction process, with ID peaks corresponding to the most brain-like representations. \citet{cheng-etal-2023-bridging} further bridges geometric and information-theoretic compression, showing that lower ID predicts faster adaptation in LMs. A few studies apply ID to practical NLP tasks. \citet{tulchinskii2024intrinsic} use ID to differentiate LLM-generated and human-written text, while \citet{yin2024characterizing} introduce local ID as a metric for hallucination detection. On the other hand, \cite{cheng-etal-2024-learning} leverages intrinsic dimension to refine word embeddings, improving model performance and explainability. However, how ID evolves across layers during reasoning and decision-making remains underexplored. Similarly, while \citet{doimo2024the} examines differences in internal geometry induced by fine-tuning vs. in-context learning, they focus on semantic clustering and representation alignment, not decision making or ID-based trends. Our work fills this gap by analyzing how ID relates to decisiveness during inference, across multiple open-weight LLMs, considering reasoning a central theme. Complementary to recent works \citep{cheng2025emergence, valeriani2023the, doimo2024the}, our study shifts focus from abstraction alone to how LLMs geometrically transition from context encoding to decision formation. 
More specifically, our study reveals specific trends across multiple models for different reasoning tasks, where the model concretizes its decision throughout layers. We find consistent geometric trends that reflect model predictions and decisiveness (see \S \ref{sec:results}). In particular, we show that ID peaks often coincide with, or slightly precede, the layer at which the model becomes most semantically committed to an answer; we validate this by projecting the mid-layer representations (resid-post and MLP-out) back to the vocabulary space. These findings provide a new geometric perspective on LLM reasoning, linking representational compression to decision formation.

%% file: sections/background.tex
\section{Internal Reperesentations in Language Models} \label{sec:background}

\noindent\textbf{Transformer-based Language Modeling:} A Language Model (LM) can be modeled as a function $f$ (parameterized by a neural network based architecture) that helps map a sequence of input tokens (prompt) to output a vector of logits, where each entry corresponds to a token in a pre-defined vocabulary. In our study, we primarily focus on the transformer-based decoder-only architectures that are trained in an autoregressive fashion, that are widely adopted by most language models 
\citep{openai2024gpt4technicalreport, geminiteam2024geminifamilyhighlycapable}. 
Given a vocabulary $\mathcal{V}$, an autoregressive language model $\mathcal{M}_\theta$ ($\theta$ denotes the model parameters) learns a parameterized function that maps an input space $\mathcal{X}$, containing a sequence of tokens $x = [x_1, \ldots, x_{t-1}] \in \mathcal{X} \subseteq \mathcal{V}^{t-1}$
 to an output probability distribution  $P_{\mathcal{M}_\theta} : \mathcal{V} \rightarrow [0, 1]$, that helps predicting the next token ($x_t$) given the sequence of previous tokens $P(x_t \mid [x_1, \ldots, x_{t-1}])$. 
Internally, the transformer-based language models consist of transformer blocks/layers ($f_{\theta_1}, f_{\theta_2}, \ldots f_{\theta_L}$) stacked together that read information from and write onto the residual stream (connected by residual connections, see \citet{elhage2021mathematical} for more details), i.e. for an input sequence $[x_1, \ldots, x_{t-1}]$, the model considers representations corresponding to each token ($x_i$) and finally predicts the distributions corresponding to the next tokens $[x_{2}, \ldots, x_{t}]$. For our study, we only consider representations corresponding to the last token in the input prompt, i.e., the token responsible for answering the query present in a prompt. 
%
After the last transformer block, the final state of the residual stream is passed
through a LayerNorm, which is further then projected onto the vocabulary space via a weight matrix $\mathbf{W}_U \in \mathbb{R}^{|\mathcal{V}| \times d_{model}}$  (also known as Unembedding layer \citep{elhage2021mathematical}).
The final probability distribution $P_{\mathcal{M}_\theta}(x_t)$ is obtained by passing the obtained logits to a softmax, leading to the final prediction. Our goal is to study the geometry of representations learned by these transformer blocks, we consider the representations corresponding to the last token for each layer’s output/transformer blocks (i.e. the output corresponding to the MLP layers present in each transformer block) that writes onto the residual stream (\S\ref{sec:experiments}) (also see Figure \ref{fig:intrinsic-diagram-thumbnail}). 
Note, unlike CNN-based vision models, the extrinsic dimensions after each layer remain the same ($\mathbb{R}^{d_{model}}$) for the transformer-based models, making the comparison between the layers more reliable.

\noindent\textbf{Intrinsic Dimension:} Intrinsic Dimension is defined as the minimum number of dimensions required to describe the data manifold with minimal information loss \citep{prml-bishop}. 
More formally, a set of (data/feature/vector) points $\mathcal{D} \subseteq \mathbb{R}^N$ are said to have intrinsic dimension (ID) equal to $d$ if its elements lie entirely, without information loss, within a $d$-dimensional manifold of $\mathbb{R}^d$, where $d < N$ \citep{CAMASTRA201626}. 
The problem of estimating the dimensions of the underlying manifold, considering a data-generating process, has been an area of interest for the last two decades \citep{MLE_id, TwoNN_id, Bac_2021-Scikit-Dimension}. 
Specifically, the DL community usually prefers estimators based on the scale of the distances between the data points due to their robustness and reliability \citep{ansuini2019intrinsic, gong2019intrinsic, pope2021intrinsic}. 
A common approach to computing intrinsic dimensions given a set of points is to investigate the space around each point and assume a constant density within the local neighborhoods; the data generation process can be modeled using a homogeneous Poisson Point Process (PPP) \citep{streit2010poisson}. For our setup, we consider three ID estimators (App. Fig. \ref{fig:intrinsic-estimators} summarizes the estimators):

\noindent\textbf{MLE} \citep{MLE_id}:
Assuming the data generation processing as a PPP, the MLE estimator formulates a likelihood expression as a function of the local intrinsic dimension specific to a data point $x$, resulting in the following maximum likelihood estimate 
$$
\hat d_k(x) = \left(\frac{1}{k-1} \sum_{j=1}^{k-1} \log \frac{T_k(x)}{ T_j(x)}\right)
$$
where $T_i(x)$ represents the distance of the $i^{th}$ nearest neighbor from the data point $x$, and $k$ is a hyperparameter for the estimator, making MLE a local intrinsic dimension estimator, i.e., it estimates the intrinsic dimension in the neighborhood of a particular point.
For calculating the global intrinsic dimension of the datasets, these local dimensions are aggregated using either the arithmetic or the harmonic mean operator \citep{mackayCorrection}:
$$
\hat d = {\frac{1}{|\mathcal D|}}\sum_{x\in \mathcal D} \hat d_k(x) \quad \text{OR} \quad \hat d = \left({{{\frac{1}{|\mathcal D|}}\sum_{x\in \mathcal D} \frac{1}{\hat d_k(x)}}}\right)^{-1}
$$
\noindent\textbf{TwoNN} \citep{TwoNN_id}:
is a global intrinsic dimension estimator that builds upon the same assumption of points coming from a PPP and formulates ID estimate using a relationship between the cumulative distribution of the random variable $\mu$, defined as the ratio of the distance between the second and the first nearest neighbor and the intrinsic dimension of the dataset and prove it to be Pareto distributed, i.e., $\mu = \frac{T_2(x)}{ T_1(x)} \sim \mathtt{Pareto}(1,d)$, 
$$
d = - \frac{\log (1 - F(\mu))}{ \log \mu}
$$
where $d$, is the intrinsic dimension, $F(\mu)$ is the cumulative density function. The real-world datasets being i.i.d, the cumulative density function can be estimated given a set of data points as 
$$
F_\mathrm{emp}(x) = \frac{1}{ |\mathcal D|}\sum_{y\in\mathcal D}{\mathbb I\{\mu(y)} \le \mu(x)\} 
$$
\noindent\textbf{GRIDE} \citep{GRIDE_id}: is a recent work that mitigates the sensitivity of TwoNN towards noisy datasets, and provides a generalization over TwoNN. 
Instead of taking the ratio of the second and the first nearest neighbor, GRIDE uses higher order distances, i.e., the ratio between the $n_2^{th}$ and the $n_1^{th}$ nearest neighbor, where $n_1$ and $n_2$ are the hyperparameters of the estimator, i.e., $\mu = \frac{T_{n_2}(x)}{ T_{n_1}(x)}$ which is distributed as 
$$
f(\mu) = \frac{d(\mu^d - 1)^{n_2-n_1-1} }{\mu^{(n_2-1)d + 1}\beta(n_2-n_1,n_1)}
$$
where $\beta(\cdot, \cdot)$ is the beta function. The log-likelihood is maximized for the above distribution, resulting in the following optimization problem in $d$, 
\begin{multline*}
\max_{d} \log \mathcal L(d) \equiv \max_{d} (n_2-n_1+1)\sum_{i=1}^{|\mathcal D|} \log (\mu_i^d -1)
+|\mathcal D|\log d -(n_2-1)d\sum_{i=1}^{|\mathcal D|}\log \mu_i
\end{multline*}
where $\mu_i $ is the ratio of distances corresponding to the $i^{th}$ data point and $d$ is the intrinsic dimension, leading to the above concave optimization problem
(see App. \ref{app:id-estimators}).
Some of the other recent extensions include Hidalgo \citep{allegra-2020-hidalgo}, which is more robust when the generated datasets have multiple underlying manifolds; we leave this for future analysis and assume representations coming from a single manifold for this study.

%% file: sections/experiments.tex
\section{Experimental Setup} \label{sec:experiments}

\noindent\textbf{Decisiveness through Multiple-Choice Prompting} In this work, we stick to reasoning captured using multiple-choice question answering (MCQA)-style prompts \citep{robinson2023leveraging, Wiegreffe2024AnswerAA}. The MCQA setup provides a principled and constrained setting for investigating the internal decision-making processes of LLMs. Unlike open-ended or cloze-style generation, MCQA structures the task as a selection among discrete alternatives, thereby reducing confounding factors related to token frequency, length bias, and linguistic fluency \protect\citep{incontextfewshotlearners}. This format enables precise analysis of the transition from contextual representation to decision, making it well-suited for studying the geometry and structure of intermediate representations. Prior work in mechanistic interpretability has focused on identifying circuits and submodules responsible for reasoning \citep{elhage2021mathematical,olsson2022incontextlearninginductionheads}, while probing studies have examined layer-wise information flow \citep{tenney-etal-2019-bert,hewitt-manning-2019-structural}. However, relatively little attention has been given to how these representations evolve geometrically to support discrete reasoning tasks. By leveraging MCQA, we aim to isolate and examine the structural properties (specifically intrinsic dimensions) of hidden states as they converge towards a decision, providing insight into the representational dynamics that govern model predictions. In our case, each input prompt is composed of:
\textbf{1) query information} ($query$): which includes the information related to that specific instance of the dataset. \textbf{2) A Choice Set} 
(\textbf{A.}\  $o_{correct}; \ \textbf{B.}\ o_{wrong}$) consisting of two or more options from which the LLM must select the correct answer and generate as output the correct choice text: \textbf{A} or \textbf{B}, or \textbf{C}, etc. \citep{robinson2023leveraging,Wiegreffe2024AnswerAA,joshi2025calibration,joshi-etal-2025-towards}.
Note that the \textbf{A.} and \textbf{B.} are for representation, and in the actual run, the correct/wrong options are shuffled to marginalize the effect of models choosing a specific option.
The prediction by the LLM ($\mathcal{M}_\theta$) depends on the above two critical components.
Additionally, the predictions also depend on how the query is framed, i.e., the prompt template ($x_{\epsilon}$) used to frame the queries.
The predicted probability/logit value of the next token can be written as:
%
%
$$
P(x_t|x_{i<t}, \mathcal{M}_{\theta}) = P(x_t|x_{query} , x_{options} , x_{\epsilon}, \mathcal{M}_{\theta})
\quad
\text{;}
\quad
\mathcal{M}_{\theta} = \{f_{\theta_1}, f_{\theta_2}, \ldots f_{\theta_L}\}
$$
$$
x_{query} \leftarrow s_i \sim \mathcal{D}
\quad 
\text{;} 
\quad 
x_{options} \leftarrow \{
\text{\textbf{A. }}
o_{correct}, 
\text{\textbf{B. }}
o_{wrong}\}
\text{;}
\quad
x_{\epsilon} \in \text{set of prompt  templates}
$$

%
where $s_i$ is a sample/instance from the language-based dataset $\mathcal{D} := \{s_1, s_2, \ldots, s_N\}$ of size $N$.
In the LLM the input prompt ($x_{i<t}$) is passed through a sequence of transformer blocks/layers ($f_{\theta_1}, f_{\theta_2}, \ldots f_{\theta_L}$), providing a distribution of logits over the vocabulary for the next tokens ($x_1, x_2, \ldots, x_t$), we only consider the predicted distribution of the last token ($x_t$), i.e., the token responsible for predicting the next plausible token or answering the question query: $\mathcal{M}_\theta(x_{i<t}) = f_{\theta_L}(\mathtt{I} + f_{\theta_{L-1}}( \ldots (\mathtt{I} + f_{\theta_1}(x_{i<t})))$. 
%
%
These sequences of operations play a crucial role in modifying the residual stream {(the $\mathtt{I} +$ denotes the update in the residual stream)}, leading to the final predicted token $x_t$. Essentially, the model $\mathcal{M}_\theta$ processes the input and predicts the next token, which is expected to be the correct option identifier (e.g., "\_A", "\_B", etc.). This token serves as a clear decision point, providing a precise locus for analyzing representational geometry.

\noindent\textbf{Representation Extraction and Decision Emergence}
As the prompt flows through the transformer layers, we collect intermediate hidden representations ($h_j$) corresponding to the final (decision) token position after each layer $j \in [1, L]$. These vectors trace how the model updates its beliefs through residual stream modifications. Mathematically: $h_j = f_j(x_{i<t}) = f_{\theta_j}(\mathtt{I} + f_{\theta_{j-1}}( \ldots (\mathtt{I} + f_{\theta_1}(x_{i<t}))) $.  
%
The representations correspond to the MLP module of the transformer block that writes back to the residual stream, i.e., $f_{\theta_j}$ represents the operations in the $j^{\text{th}}$ transformer block. We extract the representations from two places in the transformer block: \textbf{1) the MLP component} (i.e., after the nonlinearity, before writing back to the residual stream), and \textbf{2) the Resid-Post} (i.e., after the residual stream is updated/written by adding the MLP representations) corresponding to the final token (see Fig. \ref{fig:intrinsic-diagram-thumbnail}). 
These per-layer activations, collected across a dataset $\mathcal{D} = \{s_1, \ldots, s_N\}$, define the manifold structure from which we compute ID using standard estimators: MLE, TwoNN, and GRIDE. Similar to \cite{cheng2025emergence}, we choose representations corresponding to the final token for computing the intrinsic dimension, as this token represents the model’s predicted answer and is expected to encapsulate all information necessary for the prediction. Given a dataset $\mathcal{D}$, we get a space of these representations for each layer’s output corresponding to text instances present in the dataset, forming a set of $|\mathcal D|$ features/representations of the underlying manifold. Note that we only consider the representations corresponding to the last token to form this set. We make use of Transformer-Lens \citep{nanda2022transformerlens} for saving the corresponding representations. The obtained set of vectors is further considered to estimate the ID of the underlying manifold formed by these transformer blocks. Note, in actual transformer implementation, there are two points in a single transformer block where the computational blocks read/write back from/to the residual stream (self-attention and MLP); we skip the mid-skip connection in the equations above for brevity.

\begin{figure*}[t]
\begin{center}
\centerline{\includegraphics[width=\textwidth]{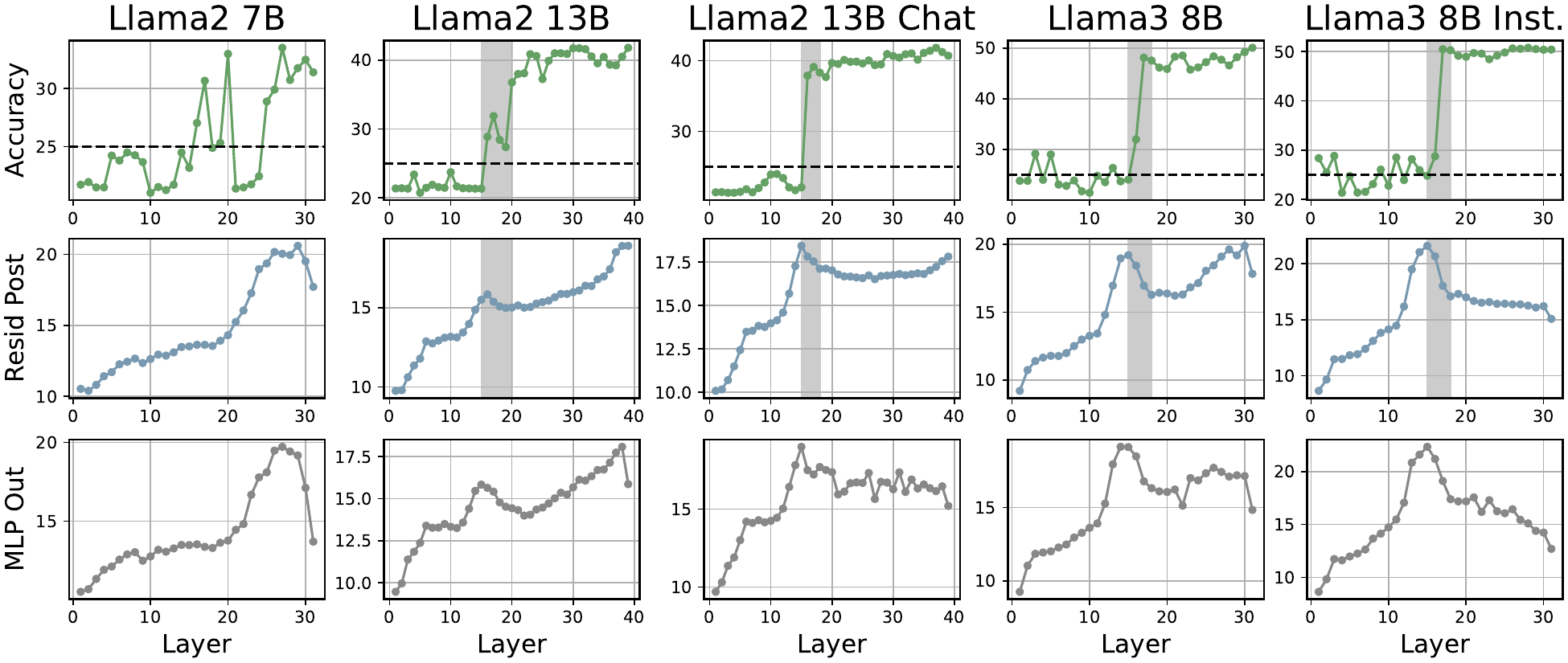}}
\vskip -0.1in
\caption{Accuracy along with ID trends for LLaMA model variants on the MMLU STEM dataset.}
\label{fig:peak-trend-MMLUSTEM}
\end{center}
\end{figure*}

\begin{figure*}[t]
\begin{center}
\centerline{\includegraphics[width=\textwidth]{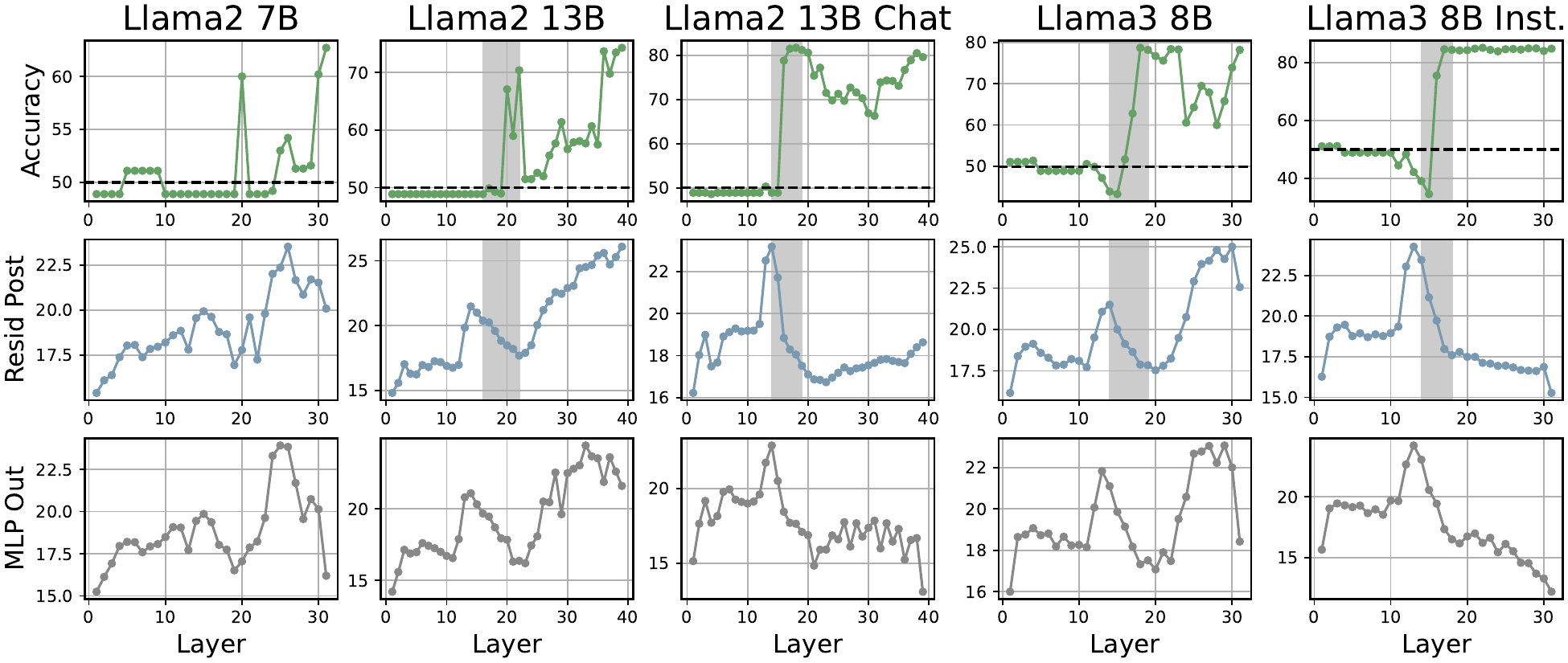}}
\vskip -0.1in
\caption{Accuracy along with ID trends for for LLaMA model variants on the COPA dataset.}
\label{fig:peak-trend-COPA}
\end{center}
\vskip -0.35in
\end{figure*}

\noindent\textbf{Measuring Representation Quality via Logit Accuracy}
To quantify the semantic sharpness of each layer’s representation, we take inspiration from Logit Lens \citep{logit-lens, haviv2023understandingtransformermemorizationrecall}, and compute the accuracy of representations at different layers (see Fig. \ref{fig:intrinsic-diagram-thumbnail}). Considering the stacked set of transformer blocks, writing sequentially over the residual stream, we take the representations after each transformer block (MLP and Residual) and project it to the vocabulary space by multiplying it with the unembedding  matrix directly ($\mathbf{W}_U$), i.e., $\mathtt{logits}(z_t) = \mathbf{W}_U\mathtt{LayerNorm}(z_t)$,  
where $z_t$ is the representation corresponding to the last token. 
{We report accuracy using the Residual Post representations rather than the MLP outputs, as the residual stream carries the accumulated state that the model propagates forward across transformer blocks. In contrast, the MLP output contains only the delta/difference added to the residual stream, representing a more localized, high-leverage adjustment rather than the full signal that contain the context. Consequently, the Residual Post signal provides a more presentable view of the model’s evolving decision state.
Note that we focus on the representation corresponding to the last token, since in autoregressive transformers, this position uniquely has access to the entire context and is solely responsible for generating the next-token prediction. From a decision-making perspective, this is the point at which the model must commit to an output, making it the most informative location for understanding decision-making from a representational perspective. While analyzing intermediate tokens could yield complementary insights into how information is distributed and refined, the last-token view most directly captures where and when the model’s decision solidifies.}
We consider the obtained logits to compute the accuracy of the representations corresponding to a particular layer. The obtained performance estimates not only help quantify the quality of representations but also provide the localization in layers where the model starts to be decisive about the decision/answer/next token. The token-level accuracy at each layer conveys how often that representation alone predicts the correct answer. This metric, coupled with ID, lets us localize the decision emergence layer: the point in the network where the model becomes sharply predictive and the representation starts collapsing into a low-dimensional manifold.
In Figures \ref{fig:peak-trend-MMLUSTEM} and \ref{fig:peak-trend-COPA} (see App. Table \ref{app-tab:plots-ref-table} for other datasets), we visualize this phenomenon across layers and models. We consistently observe that accuracy peaks and ID drops at the same layer, suggesting a tight coupling between task certainty and representational compactness.

\noindent\textbf{Real-World Tasks}
We first examine LLM behavior on real-world, language-based tasks (MCQA format) where generalization is the key. 
We specifically choose tasks (and corresponding datasets) related to linguistic abilities (Dataset: CoLA), topic knowledge (Dataset: AG News), field-specific knowledge (MMLU: STEM, humanities, social sciences, other), sentiment analysis (Rotten Tomatoes, SST2), and reasoning abilities (Causal reasoning: COPA, COLD). Note that synthetic datasets (template-based) also capture mathematical reasoning in some sense; however, they cannot be considered real-world due to template-based generation. 
We make minor changes to the template part of the prompt query ($x_{\epsilon}$) for each dataset. We provide details of datasets in the App. \ref{app:sec-real-dataset}. We use different prompt templates (details in App. \ref{app:prompt-templates}) for different datasets with minimal changes, keeping the MCQA format consistent. We experiment with various models like Llama2 family, GPT-2 family, Mistral, Phi family, and Gemma family (details in App. \ref{app:open-weight-models}). 

\noindent\textbf{Synthetic Tasks: Controlled Learning Trajectories}
Another aspect of LLMs is their open-ended reasoning via generation. 
To complement our real-world findings on MCQA-based reasoning, we analyze simplified template-based reasoning tasks where we can observe LLMs learning from scratch under tightly controlled conditions. 
To monitor the improvements throughout the training trajectory, we require reasoning datasets with low complexity for a comparison that could work for both smaller as well as larger models. We choose the Greater Than (GT) task introduced by \citet{hanna2023how-greater-than} for simplicity and the arithmetic task \citep{razeghi-etal-2022-impact-arithmetic-dataset}, considering its usage by the Pythia suite to monitor performance during model training. \textbf{The Greater Than (GT) task} consists of examples of the format “The war lasted from the year 1743 to 17 $\xrightarrow{} xy$”, where the language modeling objective is to assign a greater probability to continuations $44, 45, \ldots, 99$ than $00, 01, \ldots , 42$. 
Random accuracy is upper bounded by $99/|\mathcal{V}|$, where $|\mathcal{V}|$ is the vocab size. \textbf{The arithmetic task} also follows a template that consists of input operands $x_1 \in [0, 99]$ and
$x_2 \in [1, 50]$ and an output $y$, i.e. “Q:What is $x_1 \# x_2$? A:” with $\#$ being “plus” for addition and “times” for multiplication. We measure the accuracy of a prompt instance by checking the model’s prediction against the label y, making the random accuracy $1/ |\mathcal{V}|$. These tasks provide low input complexity but require abstract reasoning, making them ideal for analyzing how internal manifolds evolve over time. We use the Pythia model suite \citep{biderman2023pythia}, a family of 16 autoregressive transformers (14M–6.9B parameters) (see Fig. \ref{app-fig:arithmetic-peak-trend}, \ref{app-fig:greater-than-peak-trend}). With 154 publicly released training checkpoints per model, we can track the formation of decision-critical layers from early to late training (details in App. \ref{app:open-weight-models}).

%% file: sections/results.tex
\section{Results and Trends} \label{sec:results}

We conduct a large-scale empirical study analyzing the ID of representations across multiple LLM architectures, tasks, and input prompting settings. Due to space constraints, we describe the main results here, and the remaining ones are provided in the Appendix. 

\noindent\textbf{Layerwise Geometry and Task-Specific Trends} Fig. \ref{fig:peak-trend-MMLUSTEM} and \ref{fig:peak-trend-COPA} reveal how ID and accuracy evolve across the transformer layers for MMLU-STEM and COPA, respectively. Across models, we observe a characteristic “hunchback” shape in the MLP output's ID profile, i.e. ID increases in early layers, peaks at a mid-network depth, and then declines. Notably, this geometry emerges only in settings where model accuracy rises significantly above baseline (dashed lines), indicating that the presence of the hump is a marker of non-trivial abstraction and task-specific decision-making.
The residual post-activations, in contrast, display smoother and more monotonic ID changes, consistent with their role in progressively aggregating signals across layers. This distinction is especially prominent in the contrast between reasoning and retrieval tasks. In COPA (Fig. \ref{fig:peak-trend-COPA}), models must synthesize causal and contextual information; here, the ID peak is sharp, and the post-peak ID drop coincides with decisive increases in prediction accuracy. In MMLU-STEM (Fig. \ref{fig:peak-trend-MMLUSTEM}), a fact-retrieval task, the ID trend is flatter, and the accuracy increases monotonically across layers with no prominent compression phase. Interestingly, we observe a striking alignment between ID peaks and abrupt accuracy shifts in MMLU; the sharpest increase in accuracy always follows the ID peak. These observations suggest that transformer layers undergo an information compression phase just prior to forming confident predictions, supporting the idea that compression marks the onset of semantic decisiveness.
{While these patterns reveal a strong correlation between geometric compression and model decisiveness, we emphasize that the relationship is not strictly causal. The alignment of ID peaks with an increase in accuracy suggests that representational geometry and decision confidence co-evolve, with compression emerging just before the model commits to a prediction. This ordering provides a weak hint at an underlying causal structure that is worth investigating further; however, at present, the evidence should be interpreted as correlational rather than causal.}

\noindent\textbf{MLP Outputs vs. Residual Post-Activations} 
{
A key finding is the distinct behavior of MLP outputs compared to residual post-activations (as shown in the zoomed-in version in App. Figure~\ref{fig:resid_post_vs_mlp_out}). 
While residuals reflect a smoothed integration of signals across the network, MLP outputs consistently display sharper ID transitions, highlighting their role in injecting task-specific refinements. 
In other words, the residual stream represents the model’s continuously updated internal state, an accumulated integration of information that is propagated across layers. In contrast, the MLP output reflects a targeted modification to this stream, often acting as a high-leverage “correction” that sharpens or reorients the representation toward task-relevant directions.
Consequently, the ID trajectories of these two signals reveal complementary aspects of computation, residual post-activations evolve smoothly, capturing the gradual stabilization of meaning, while MLP outputs exhibit sharper, more localized ID transitions, corresponding to points of semantic refinement or decision formation.}



\begin{figure*}[t]
\begin{center}
\centerline{\includegraphics[width=1\textwidth]{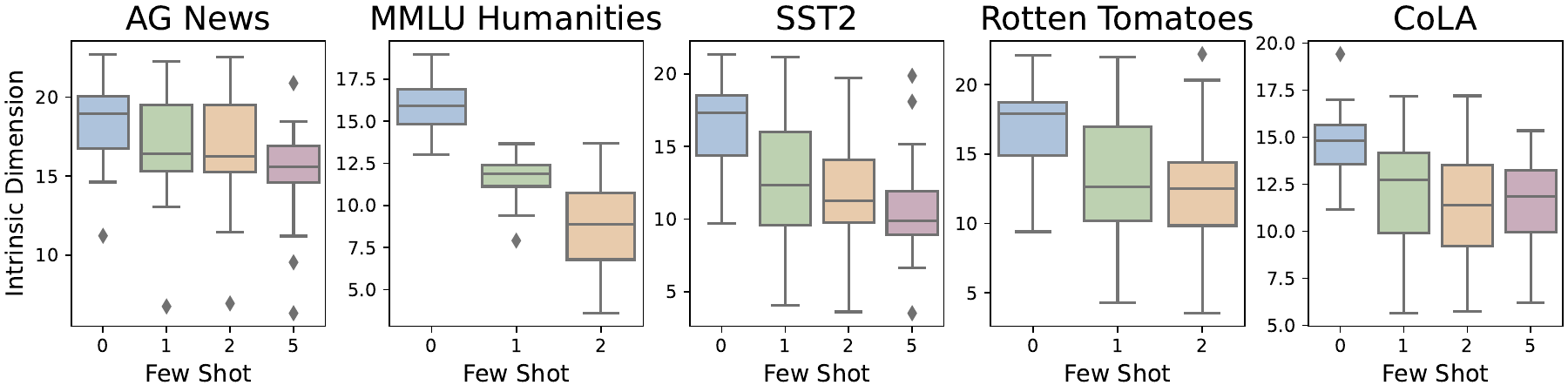}}
\caption{The figure shows the ID of the last layer (MLP Out) feature representation in the in-context learning setting. The box plot shows the distribution of ID for all the \openweightmodelcount open-weight models. Overall, we observe IDs decreasing as more number of examples are provided in the context.}
\label{fig:icl few shot intrinsic}
\end{center}
\end{figure*}


\noindent\textbf{Few-Shot Prompting Accelerates Compression} 
Few-shot prompting modulates model geometry in systematic ways. Fig.  \ref{fig:icl few shot intrinsic} and \ref{app:fig:hunchback-rottentom} show that increasing the number of in-context examples lowers the final-layer intrinsic dimension, especially in MLP outputs, indicating that few-shot prompts induce more efficient compression of the input space. This pattern is particularly salient in reasoning-heavy tasks, where compression accelerates with each additional example, suggesting that LLMs generalize better when they can abstract patterns across shots. Moreover, as shown in App. Fig. \ref{app:fig:hunchback-rottentom} and App. Figs. \ref{app-fig:hunchback-MLE}–\ref{app-fig:hunchback-Gride}, well-performing models exhibit earlier ID peaks and steeper ID declines, reinforcing the view that efficient compression precedes confident prediction. 
We also observe that across datasets, models of varying sizes follow similar normalized ID trajectories when aligned by relative model depth (from 0 to 1). As shown in App. Figs. \ref{app-fig:interpolated-correlation-agnews}–\ref{app-fig:interpolated-correlation-mmlu-other} (see App. Table \ref{app-tab:plots-ref-table}), the ID profiles maintain high inter-model correlation, highlighting that despite architectural variation, models learn consistent representational transformations. This suggests that LLMs may converge towards a shared geometric inductive bias when trained on language data.
\begin{figure*}[t]
\begin{center}
\centerline{\includegraphics[width=\textwidth]{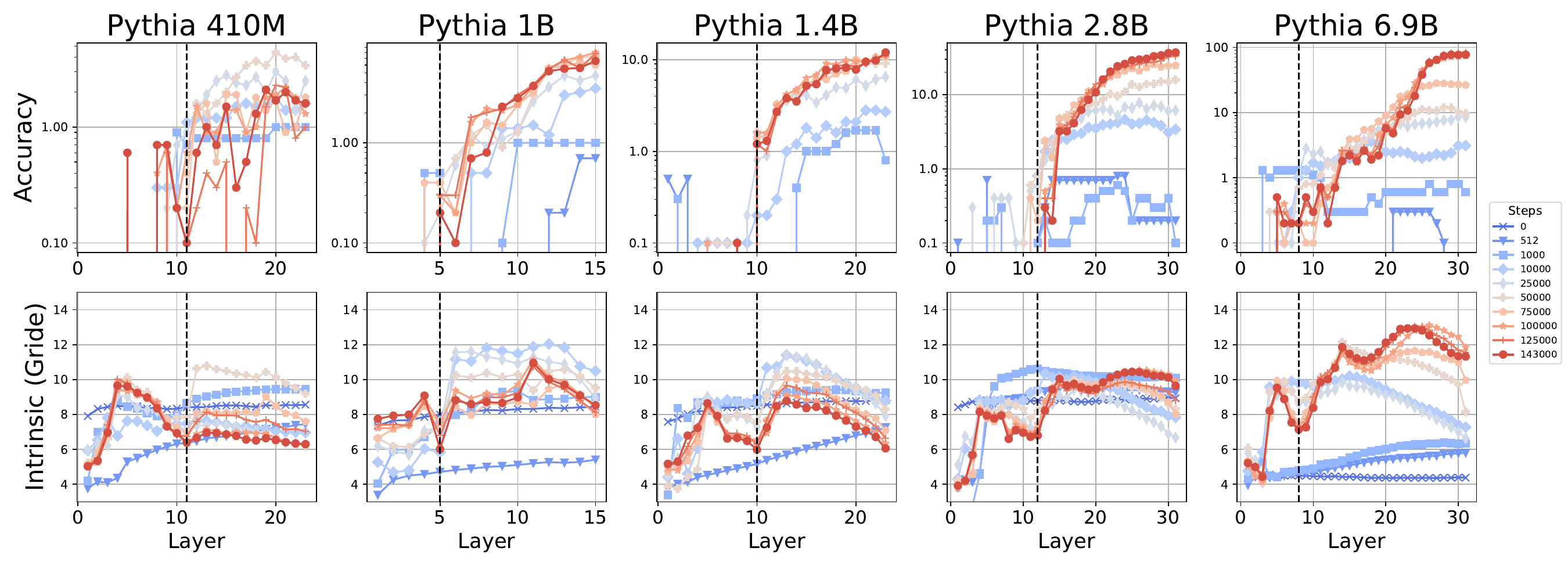}}
\vskip -0.1in
\caption{
ID of residual post hidden layers in Pythia series models evolving throughout training for the Arithmetic dataset. The red curve shows the final checkpoint for architectures of different sizes. The top row shows the quality of layer representations in the form of accuracy (log scale). Interestingly, we observe that the model starts to be decisive about the correct token, where the ID shows a reverse peak (highlighted as black dashed vertical lines). 
}
\label{fig:peak-trend}
\end{center}
\end{figure*}

\noindent\textbf{Predictive Utility of ID performance} 
We observe that final-layer ID values negatively correlate with model accuracy across multiple datasets in the LLaMA family (App. Table \ref{tab:llama-correlation-results}) but the magnitude of the correlation varies a lot for different datasets, hence, the final-layer ID estimates corresponding to the last token offers a very weak unsupervised and architecture-agnostic proxy for model generalization. We also observed that when all the models that perform better than the baseline accuracy are considered, the correlation disappears (App. Table \ref{tab:correlation-all-models}). 
In contrast to results reported by prior arts \cite{ansuini2019intrinsic,pope2021intrinsic,birdal2021intrinsicdimensionpersistenthomology}, we observe that the last layer ID (showing weak correlation) can not always be used as a strong proxy for accuracy/error across tasks.

\noindent\textbf{Understanding Training Dynamics via Synthetic Tasks} 
To better understand how intrinsic dimension (ID) evolves during learning, we turn to the Pythia model family (also see App. \ref{app:open-weight-models}). Fig. \ref{fig:peak-trend} tracks ID estimates and accuracy across layers and checkpoints. 
During training, we observe the emergence of a hunchback-like trend in ID, i.e., first increasing as representations diversify during early training, then peaking mid-network, and eventually decreasing toward the output layers.  Interestingly, the relationship between ID and accuracy diverges from what we observe in MCQA tasks. In the higher-capacity models, accuracy begins to rise immediately after ID reaches a low (reverse peak). However, unlike MCQA tasks, where decisive ID transitions are tightly aligned with accuracy jumps, the transitions in generative reasoning tasks are more gradual, suggesting a more continuous integration of symbolic structure. These findings highlight two key insights. First, ID evolution during training can reveal whether a model is generalizing or merely memorizing, providing a geometric lens into the learning process. Second, the ID-based compression trends observed in real-world MCQA tasks are not artifacts of option-based formats; they also emerge, although more gradually, in open-ended generative reasoning tasks, especially in models with sufficient capacity. Thus, ID provides a unified view of learning geometry that applies both during training and across diverse reasoning paradigms.

\noindent\textbf{Scaling In-Context Learning: Geometry of Few-Shot Adaptation}
To study how transformer representations evolve under extreme in-context learning (ICL) conditions, we analyze up to 50-shot prompting in arithmetic reasoning tasks using Pythia models of varying sizes (410M to 6.9B) (see App. Fig. \ref{fig:few-shot-pythia}). Interestingly, we found that the accuracy decreases after some examples with smaller models; this could be due to the fact that the model starts extracting surface patterns instead of generalizing and predicting the output according to the inherent arithmetic operation. Overall, we find that increasing the number of few-shot examples consistently improves accuracy while reducing the ID of final-layer representations, especially in larger models (2.8B being an exception). 
These trends highlight a tight coupling between few-shot generalization and latent space compression, as models condition on more examples, they restructure their internal geometry to form more compact, decision-relevant manifolds. Notably, this is the first study to probe up to 50-shot prompting in this context, positioning intrinsic dimension as a promising unsupervised proxy for evaluating ICL efficiency and saturation in LLMs.

Overall, the ID often relies on a smaller range of (5, 37) when compared to the extrinsic dimensions (aka model hidden dimensions (768, 4096)), irrespective of size and number of layers present in these models. We found this trend to be consistent for both template-based synthetic datasets as well as real-world datasets, pointing toward the language-specific tasks being present in low-dimensional manifolds. 
We provide additional results, discussion, and future directions in the App. \ref{app:additional-results-section}.  

%% file: sections/limitations.tex
\noindent\textbf{Limitations}
Though our work considers a wide range of open-weight models along with synthetic as well as real-world datasets, there is still room for experimentation with more language data sources. In our work, we primarily considered a setting where only a token is used for prediction (computing performance and intrinsic dimensions) and not the generative modeling setting, where multiple tokens are generated in an open-ended autoregressive fashion. Extending this analysis to Natural Language Generation (NLG) tasks becomes difficult due to the inherent autoregressive nature of these models. Another major limitation comes from the ID estimates that we use. In general, though prior arts have considered them for estimating ID estimates of features, these estimators often provide a noisy ID estimation of the underlying manifold, providing only a weaker estimate. Though we make use of a recently improved ID estimator (GRIDE), there still remains some scope for improvement. In the future, it would be interesting to revalidate these estimations via more advanced ID estimators.  Moreover, in this work, the primary focus was to observe the trends across a wide range of models, and we only considered the features transformed by the transformer blocks for analysis, leaving the hidden representation inside these models aside. However, the proposed experimental setup could be utilized to study the spaces learned by each submodule of the transformer blocks (MLP-heads/Attention/LayerNorm/etc). 

Further, on a broader level, exploring the relationship between intrinsic dimension and entropy provides a promising bridge between geometric and information-theoretic perspectives. Recent studies (e.g., \citealp{skean2025layer, Stolfo}) have tried linking entropy to decision making, but examining this connection throughout the network could reveal how decision-making and representational geometry co-evolve. This line of work may ultimately lead to a unified framework where the activation geometry helps characterize the emergence of reasoning and understanding decision-making across layers.


%% file: sections/conclusion.tex
\section{Conclusion}
In this work, we find that LLMs (stacked transformer layers) project the datasets into low-dimensional manifolds with ID estimates considerably lower than the actual latent dimensions. 
With a detailed analysis of \openweightmodelcount open-weight models, we find that the ID peaks are strongly coupled with decision-making happening inside models.
This coupling between representational compression and model decisiveness provides geometric evidence of how abstract reasoning and prediction confidence co-evolve within the transformer architecture.
We believe this study will open up new avenues for research in understanding the low-dimensional manifolds learned by the Language models. 
More broadly, we hope this study encourages the development of geometric interpretability tools that move beyond surface-level investigation, moving toward a deeper understanding of the internal topologies that support reasoning/decision-making in LLMs. By viewing activations through the dual lenses of geometry and information, we can begin to map not just what language models know, but how their internal structure gives rise to understanding and decision formation.


\section*{Acknowledgments}

We would like to thank the anonymous reviewers and the meta-reviewer for their insightful comments and suggestions. 
This research work was partially supported by the Research-I Foundation of the Department of CSE at IIT Kanpur.

%% file: sections/appendix.tex
\section*{Appendix}

\appendix


\hypersetup{linkcolor=blue}


\startcontents[appendix] 
\section*{Table of Contents} 
\printcontents[appendix]{section}{0}{\setcounter{tocdepth}{4}} 

\startlist[appendix]{lot} 
\section*{List of Tables} 
\printlist[appendix]{lot}{}{\setcounter{tocdepth}{2}} 

\startlist[appendix]{lof} 
\section*{List of Figures} 
\printlist[appendix]{lof}{}{\setcounter{tocdepth}{2}}

\newpage

\section{Estimating Intrinsic Dimensions} 
\label{app:id-estimators}
In higher dimensions, the datasets often lie on 
on a low dimensional manifold $\mathcal Y$ and there exists a mapping $f:\mathcal Y\to\mathcal X$ where $\mathcal X$ is the data space of higher dimensions. Typically, it is considered that the function $f$ is smooth and continuous, which ensures that nearby points in the lower-dimensional manifold will also be close by in the higher-dimensional space.
In the past, multiple methods have been proposed to estimate the dimensions of the underlying low-dimensional mapping, including methods like projection or geometric-based methods like PCA \cite{fan2010intrinsic} and CPCA \cite{packingIntrinsic2002}. 
In the DL community, the nearest neighbor-based approaches are widely accepted due to their easier scalability and robust estimates. One such approach introduced by \citet{MLE_id} proposes 
a local intrinsic dimension estimator, i.e., IDs are estimated pertaining to each point present in the manifold by looking at the k-nearest neighbors. The proposed estimator assumes that in a neighborhood of a data point, $x$, the density of the data distribution is approximately constant. In this region, one can model the probability of finding another sample using a homogeneous Poisson process, which helps formulate a maximum likelihood objective in terms of the intrinsic dimension, having a closed-form solution
$$
\hat d_R(x) = \left(\frac{1}{N(R,x)} \sum_{j=1}^{N(R,x)} \log \frac{R}{T_j(x)}\right)
$$ 
where $N(R,x)$ represents the number of sampled data points found in the neighborhood of radius $R$. 

For real-world datasets, the approximate solution to the maximum likelihood problem can be stated as 
$$
\hat d_k(x) = \left({1\over k-1} \sum_{j=1}^{k-1} \log {T_k(x)\over T_j(x)}\right)
$$ 
where $T_k(\cdot):\mathcal X\to \mathbb R$ computes the distance of the $k^{th}$ closed neighbor. This approximation provides a local dimension estimator corresponding to each datapoint. 
Further, for computing the global estimate, \citet{MLE_id} propose a straightforward averaging over these values, i.e., 
$$
\hat d = {1\over |\mathcal D|}\sum_{x\in \mathcal D} \hat d_k(x)
$$
\citet{mackayCorrection} further modifies/improves the aggregation of local intrinsic dimension estimates by formulating another maximum likelihood problem, similar to the previous one, resulting in the global intrinsic dimension being the inverse of the average of the inverse of the local intrinsic dimension, i.e., 
$$
\hat d = \left({1\over |\mathcal D|}\sum_{x\in \mathcal D} {1\over\hat d_k(x)}\right)^{-1}
$$
instead of direct averaging. It is suggested to do a direct averaging over the parameter $k$ by \cite{MLE_id} for real-world datasets to compute the IDs empirically. 

TwoNN \cite{TwoNN_id} is a global intrinsic dimension estimator that uses the same assumptions as the MLE of a homogeneous Poisson process.  
However, it relies on 
the terms $T_1(x)$ and $T_2(x)$ and prove that in the intrinsic space, the ratio $\mu = {T_2(x)\over T_1(x)}$ follows the distribution $\mathtt{Pareto}(1,d)$, where $d$ is the intrinsic dimension, thus, establishing the relation 
$$
d = - {\log (1 - F(\mu))\over \log \mu}
$$
For estimation of real-world datasets, the above term is approximated by first considering the cumulative distribution as 
$$
F_\mathrm{emp}(x) = {1\over |\mathcal D|}\sum_{y\in\mathcal D}{\mathbb I\{\mu(y)} \le \mu(x)\}
$$ 
where $\mu(x)$ is computed as $T_2(x)\over T_1(x)$ for each of the given datapoint. To estimate the slope, 
a linear regressor is fitted on the dataset $\{(-\log \mu(x), \log (1-F_\mathrm{emp}(x))\}_{x\in\mathcal D}$ passing through the origin. Moreover, \citet{TwoNN_id} empirically suggests 
discarding a small $\alpha < 1$ fraction of datapoints with the highest $\mu(x)$ values, resulting in better estimates for real-world datasets.

GRIDE generalizes the idea of TwoNN by keeping the same assumptions, however, modeling the $n_2^{th}$ and the $n_1^{th}$ nearest neighbors, resulting in the probability distribution
$$
f(\mu) = {d(\mu^d - 1)^{n_2-n_1-1}\over \mu^{(n_2-1)d + 1}\beta(n_2-n_1,n_1)}
$$
As the above distribution doesn't have a closed-form solution for the cumulative function, the intrinsic dimension is estimated by maximizing the following log-likelihood
\begin{multline*}
\max_{d} \log \mathcal L(d) \equiv \max_{d} (n_2-n_1+1)\sum_{x\in\mathcal D}\log (\mu(x)^d -1)+|\mathcal D|\log d -(n_2-1)d\sum_{x\in \mathcal D}\log \mu(x)
\end{multline*}
which turns out to be a concave optimization problem. The GRIDE estimates using a general form are more robust to noisy observations present in the datasets.

\begin{figure*}[t]
  \centering
   \includegraphics[width=\textwidth]{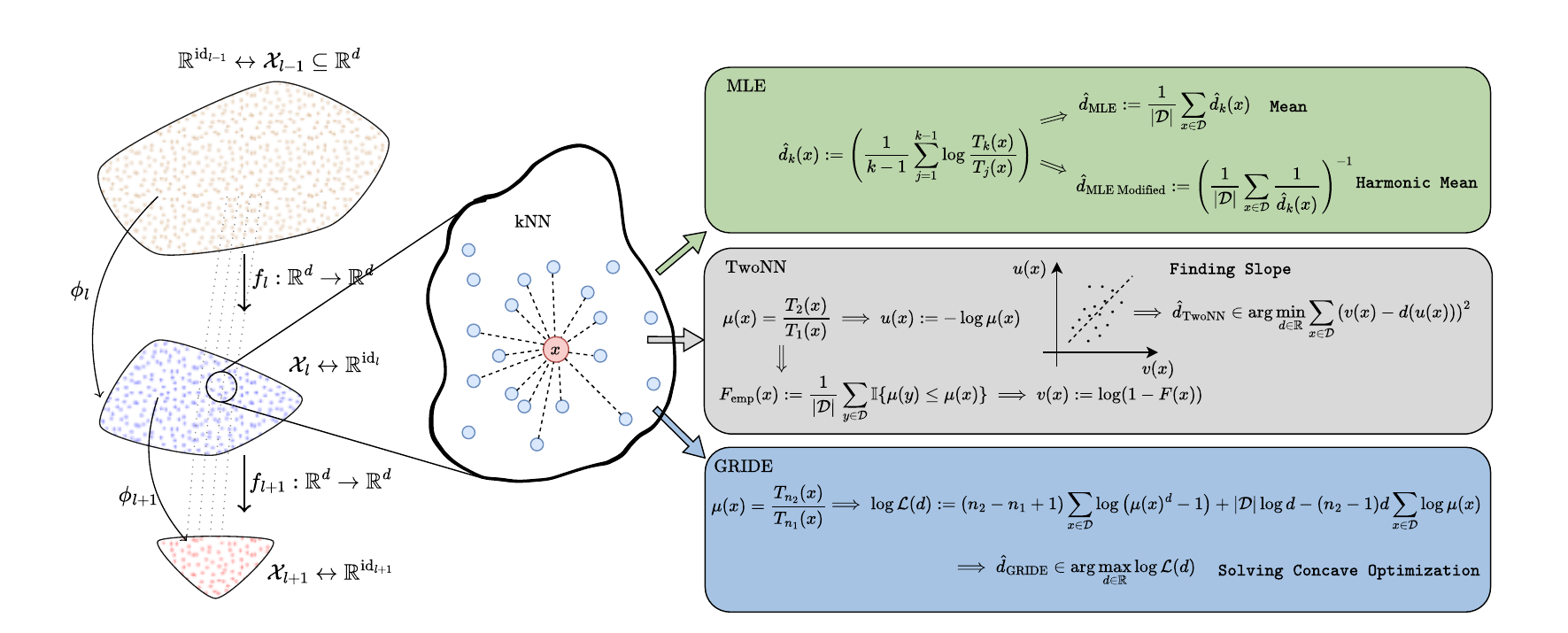}
 \caption[Overview of ID Estimators]{The figure shows an overview of the Intrinsic dimension estimators. 
 The leftmost blobs show the underlying manifolds as the features are transformed by the transformer blocks (as explained in Figure \ref{fig:intrinsic-diagram-thumbnail}). The obtained set of features is further used to compute the Intrinsic dimensions of the underlying manifold. The middle blob shows a zoomed-in version, highlighting the local neighborhood of a point $x \in \mathcal{X}_l$.
 All the ID estimators make use of local neighborhood estimates of the dimensionality. 
 MLE formulates the likelihood based on k-nearest neighbors for local ID estimation, which is further averaged using the mean or harmonic mean to compute global intrinsic dimensions. TwoNN reduces this to 2 nearest neighbors and estimates the empirical cumulative distribution $F_{emp}$ assuming the points to be i.i.d., and further uses linear regression to estimate the slope as global intrinsic dimensions. GRIDE generalizes the use of two nearest neighbors to $n_1^{th}$ and $n_2^{th}$ neighbors and frames a concave optimization problem to estimate the global intrinsic dimensions.}
 \label{fig:intrinsic-estimators}
\end{figure*}

Also, see Figure \ref{fig:intrinsic-estimators} for an overview of the different Intrinsic dimension estimators. 

\begin{table}[t]
\caption[ID estimators' hyperparameters]{The table shows the hyperparameters used for different Intrinsic dimension estimators.}
 \label{app-tab:ide-hyperparameters}
 \centering
\begin{tabular}{lcc}
\toprule
\textbf{Method} & \textbf{Parameter} & \textbf{Values}\\ 
\midrule
MLE &$k$  & all values in range [12, 24]      \\ 
MLE-Modified &$k$  & all values in range [12, 24]      \\ 
TwoNN & Discard Ratio  & 0.1           \\ 
GRIDE  & $n_1, n_2$& $20, 40$\\ 
\bottomrule
  \end{tabular}
\end{table}

\section{Details of Real World Datasets} \label{app:sec-real-dataset}
For real-world text-based tasks, we would like to investigate the overall capabilities of different sets of widely used datasets. We specifically choose linguistic abilities, topic knowledge, field-specific knowledge (STEM, humanities, other), sentiment analysis (emotional intelligence), and reasoning abilities (Causal reasoning). Note that synthetic datasets (template-based) also capture mathematical reasoning in some sense; however, they cannot be considered real-world due to template-based generation. For all these abilities, we choose specific datasets that contain text samples that help validate the task performance. We consider a common MCQA format prompt template as described in the main paper to keep the analyses comparable with each other. 
Another advantage that comes with the MCQA format is the transformation of the dataset query in a different format than the original text, making the predictions only work when the model is able to provide predictions based on generalized learned tasks and not the memorized examples. 
We make minor changes to the template part of the prompt query ($x_{\epsilon}$) for each of the datasets. 
We provide details of the datasets below:


\textbf{Linguistic}: For linguistic abilities, we consider the widely used CoLA dataset \cite{warstadt2019neural} that contains English sentences from 23 linguistics publications, expertly annotated for acceptability (grammaticality) by their original authors, making the prompts an MCQA query with 2 choices ('Accepted'/'Unaccepted') in our case.

\textbf{Topic knowledge:} For world topic knowledge, we make use of AG News dataset \cite{zhang2016characterlevelconvolutionalnetworkstext}, which contains sentences from news articles on the web, primarily covering the 4 largest classes (“World”, “Sports”, “Business”, “Sci/Tech”), making it a 4-choice MCQA query. 

\textbf{Field-specific knowledge:} MMLU \cite{hendrycks2021measuringmassivemultitasklanguage} is another widely used benchmark in the LLM evaluation/benchmarking community. The benchmark primarily aims to cover questions regarding world knowledge and problem-solving, including questions from different fields, including STEM, Humanities, Social Sciences, and Others. For each of the questions, the benchmark provides 4 choices/options. Note that for other datasets, the choices remain fixed; however, in MMLU queries, the option text is dynamic and keeps changing, making it a more complicated task for language understanding. 

\textbf{Emotional intelligence:} Affective computing is another area where language plays a vital role \cite{singh2021fine}. For this ability, we consider two known datasets, Rotten Tomatoes \cite{rottentomatoes} and SST2  \cite{socher-etal-2013-recursive}. Both of these datasets contain sentences annotated with the sentiment "Positive" or "Negative", making it an MCQA query with two choices in our setting.

\textbf{Reasoning abilities:} Real-world-based reasoning abilities are hard to capture in language benchmarks. For reasoning in real-world concepts, we found causal reasoning to be a suitable ability as it involves both real-world examples and a form of reasoning. We use COPA \cite{gordon-etal-2012-semeval}, and a small sample from the recently introduced COLD dataset \cite{joshi2024cold}. Both these datasets contain a premise event and a corresponding causal query question, along with two choices, where a system is required to predict which of the two choices is the most plausible cause/effect of the premise event. 

All these datasets cover a wide range of language understanding abilities, helping us quantify the generalization of the multiple open-weight models that we experimented with. We summarize various datasets in Table \ref{app-tab:dataset-desc}. 

\begin{table*}
\caption{The table provides details about the various real-world datasets.}
 \label{app-tab:dataset-desc}
 \centering
 \small
\renewcommand{\arraystretch}{1}
\setlength\tabcolsep{3pt}
    \begin{tabular}{lcccccc}
\toprule
\textbf{Dataset Name} & \textbf{Task Type} & \textbf{\# Samples} & \textbf{Avg. Prompt Length} & \textbf{\# Choices}\\
\midrule
AG News & Topic Knowledge & 7600 & 235.30 &  4\\
Rotten Tomatoes & Sentiment &1066  &  115.52&  2\\
SST2 & Sentiment & 872 &105.84 &  2\\
CoLA & Linguistic & 1043 &41.83 &  2\\
MMLU Stem & Field Specific Knowledge &3018  & 149.09 &  4\\
MMLU Humanities & Field Specific Knowledge &4705  & 535.10& 4\\
MMLU Social Sciences & Field Specific Knowledge &3077  & 116.35& 4\\
MMLU Others & Field Specific Knowledge &3242  & 163.32& 4\\
COPA & Causal Reasoning &1000  & 34.89 & 2\\
COLD & Causal Reasoning &1000  & 29.49 & 2\\
\bottomrule
    \end{tabular}
\end{table*}

\section{Prompt Templates}\label{app:prompt-templates}

We use different prompt templates for different datasets with minimal changes, keeping the MCQA format consistent. An input prompt given to the model helps predict the probability distribution of the next token. 
The predicted probability/logit value of the next token can be written as:
\begin{align*}
P(x_t|x_{i<t}, \mathcal{M}_{\theta}) &= P(x_t|x_{query} , x_{options} , x_{\epsilon}, \mathcal{M}_{\theta})\\
x_{query} &\leftarrow s_i \sim \mathcal{D}\\
x_{options} &\leftarrow \{
\text{\textbf{A. }}
o_{correct}, 
\text{\textbf{B. }}
o_{wrong}\}\\
x_{\epsilon} &\in \text{set of prompt  templates} \\ 
\mathcal{M}_\theta &= \{f_{\theta_1}, f_{\theta_2}, \ldots f_{\theta_L}\}
\end{align*}
where $s_i$ is a sample/instance from the language-based dataset $\mathcal{D} := \{s_1, s_2, \ldots, s_N\}$ of size $N$. We choose a general prompt template ($x_\epsilon$) for different datasets. 
Figure \ref{app-fig:general-question-prompt} shows a generalized prompt template used for all the datasets. All the datasets represent a different task, requiring a different generic query specific to the task. We modify only the generic query to make minimal changes to the prompt template.  Figure \ref{app-fig:general-question-prompt}, Table \ref{app-tab:prompt-ref} provide the references to the prompt templates used for different datasets.

\begin{table}
\caption{The table provides reference links to the prompt templates used for different real-world datasets.}
 \label{app-tab:prompt-ref}
 \centering
    \begin{tabular}{lc}
\toprule
\textbf{Dataset Name}            & \textbf{Templates (Ref.)} \\ \midrule
AGNews \cite{zhang2016characterlevelconvolutionalnetworkstext}                            & Figure \ref{app-fig:agnews_question_prompt}         \\ 
MMLU \cite{hendrycks2021measuringmassivemultitasklanguage}                              & Figure  \ref{app-fig:mmlu_question_prompt}           \\ 
CoLA \cite{warstadt2019neural}                             & Figure  \ref{app-fig:cola_question_prompt}           \\ 
RottenTomatoes \cite{rottentomatoes}                    & Figure  \ref{app-fig:rottentomatoes_question_prompt} \\ 
SST-2  \cite{socher-etal-2013-recursive}                            & 
 Figure  \ref{app-fig:sst2_question_prompt}           \\ 
COPA \cite{gordon-etal-2012-semeval}                             & Figure  \ref{app-fig:copa_question_prompt}           \\ 
COLD \cite{joshi2024cold}                              & Figure \ref{app-fig:cold_question_prompt}           \\ \bottomrule
\end{tabular}
\end{table}

\section{Details of Open Weight Models} \label{app:open-weight-models}
For our experiments, we consider a wide range of open-weight transformer-based LLMs. Specifically, we consider 
GPT-2 Small,
GPT-2 Medium,
GPT-2 Large, and
GLT-2 XL \cite{Radford2019LanguageMA} from the GPT-2 family;
GPT-Neo 125M, 
GPT-Neo 1.3B, and
GPT-Neo 2.7B, from GPT-Neo family \cite{gpt-neo};
GPT-J 6B \cite{gpt-j};
Phi 1 \cite{gunasekar2023textbooks} ,
Phi 1.5 \cite{li2023textbooksneediiphi15}, and
Phi 2 \cite{javaheripi2023phi}, from Phi family;
Gemma 2B, and
Gemma 7B from Gemma family \cite{gemmateam2024gemmaopenmodelsbased};     
Llama2 7B,
Llama2 7B Chat,
Llama2 13B, and 
Llama2 13B Chat, from Llama2 family  \cite{touvron2023llama2openfoundation};
Llama3 8B, Llama3 8B-Instruct, from Llama3 family \cite{grattafiori2024llama3herdmodels}; and   
Mistral 7B    \cite{jiang2023mistral7b}. 

All these models provide a broad spectrum of model sizes and architectural changes with different extrinsic dimensions. Table \ref{tab:Model-desc} provides the details of the used open-weight models.

\noindent\textbf{Reason for Selecting Pythia Model for Synthetic Tasks:} 
We use the Pythia model suite \citep{biderman2023pythia}, a family of 16 autoregressive transformers (14M–6.9B parameters), all trained on The Pile \citep{pile-dataset} using the same architecture, data order, and objective. With 154 publicly released training checkpoints per model, we can track the formation of decision-critical layers from early to late training. For feasibility, we analyze 10 evenly spaced checkpoints per model (steps: 0, 512, 1k, 10k, 25k, 50k, 75k, 100k, 125k, 143k), applying our ID and logit-based methods layer-wise.

To better understand how intrinsic dimension (ID) evolves during learning, we turn to the Pythia model family, which uniquely provides checkpoints at regular intervals throughout training. This allows us to directly examine the temporal dynamics of representation geometry, how model manifolds emerge, expand, and compress, as the model is exposed to more data and optimizes its objective. We focus on a synthetic arithmetic reasoning task that requires symbolic computation rather than token classification. Unlike MCQA settings where the output probability distribution is conditioned on the choices (e.g., "\_A", "\_B"), here the whole probability distribution is considered, making the task fundamentally generative.

\begin{table*}[t]
\caption[Model Description]{The table shows the list of open-weight models used for investigating the intrinsic dimensions. The list of models covers a wide range of layers with different model sizes. Note that the hidden dimension for each of the models is represented as Extrinsic Dimensions. Our experiments suggest that though these models use high extrinsic dimensions for information flow between the layers, the underlying manifold often lies in lower dimensions. }
 \label{tab:Model-desc}
 \centering
    \begin{tabular}{lccccc}
    \toprule 
        \textbf{Model} & \textbf{Size}&
        \textbf{$\#$ Layers}&
        \textbf{Layer Dimension}& 
        \textbf{Vocabulary Size}\\ 
    \midrule
GPT-2 Small 
& 85M  & 12 & 768  &50257 \\
GPT-2 Medium  
& 302M & 24 & 1024 & 50257\\
GPT-2 Large   
& 708M & 36 & 1280 &50257 \\
GLT-2 XL 
& 1.5B & 48 & 1600 &50257 \\
GPT-Neo 125M  
& 85M  & 12 & 768  & 50257 \\
GPT-Neo 1.3B 
& 1.2B & 24 & 2048 & 50257 \\
GPT-Neo 2.7B  
& 2.5B & 32 & 2560 & 50257 \\
GPT-J 6B 
& 5.6B & 28 & 4096 & 50400 \\
Phi 1  
& 1.2B & 24 & 2048 & 51200 \\
Phi 1.5    
& 1.2B & 24 & 2048 & 51200 \\
Phi 2 
& 2.5B & 32 & 2560 & 51200 \\
Gemma 2B    
& 2.1B & 18 & 2048 &256000  \\
Gemma 7B    
& 7.8B & 28 & 3072 & 256000 \\
Llama2 7B 
& 6.5B & 32 & 4096 & 32000 \\
Llama2 7B Chat  
& 6.5B & 32 & 4096 & 32000 \\
Llama2 13B    
& 13B  & 40 & 5120 & 32000 \\
Llama2 13B Chat   
& 13B  & 40 & 5120 & 32000 \\
Llama3 8B  
& 7.8B & 32 & 4096 & 128256 \\
Llama3 8B-Instruct  
& 7.8B & 32 & 4096 & 128256 \\
Mistral 7B    
& 7.8B & 32 & 4096 & 32000  \\
\midrule
Pythia 14M    
& 1.2M & 6 & 128 &50304\\
Pythia 31M                          & 4.7M & 6 & 256 &50304\\
Pythia 70M                         & 19M & 6 & 512 &50304\\
Pythia 160M                          & 85M & 12 & 768 & 50304\\
Pythia 410M                          & 302M & 24 & 1024 & 50304\\
Pythia 1B                        & 805M & 16 & 2048 &50304\\
Pythia 1.4B                          & 1,2B & 24 & 2048 & 50304\\
Pythia 2.8B                          & 2.5B & 32 & 2560 & 50304\\
Pythia 6.9B                          & 6.4B & 32 & 4096 & 50432\\

       \bottomrule
    \end{tabular}
\end{table*}

\section{Additional Results, Discussion and Future Directions} \label{app:additional-results-section}

\begin{table*}
\caption{The table shows Accuracy and corresponding last layer (MLP Out) Intrinsic Dimensions of various models on different datasets}
 \label{tab:accuracy-intrinsic-table-datasets}
 \centering
\small
\renewcommand{\arraystretch}{1}
\setlength\tabcolsep{3pt}
\begin{tabular}{lcccccccccc}
\toprule
\textbf{Model} & \multicolumn{2}{c}{\textbf{AG News}} & \multicolumn{2}{c}{\textbf{COPA}} & \multicolumn{2}{c}{\textbf{Rotten Tomatoes}} & \multicolumn{2}{c}{\textbf{SST2}} & \multicolumn{2}{c}{\textbf{CoLA}} \\
\cmidrule(lr){2-3}\cmidrule(lr){4-5}\cmidrule(lr){6-7}\cmidrule(lr){8-9}\cmidrule(lr){10-11}
 & \textbf{Acc} & \textbf{ID} & \textbf{Acc} & \textbf{ID} & \textbf{Acc} & \textbf{ID} & \textbf{Acc} & \textbf{ID} & \textbf{Acc} & \textbf{ID} \\
\midrule
Random Baseline & 25 & - & 50 & - & 50 & - & 50 & - & 70 & - \\
GPT-2 & 25.28 & 18.97 & 48.90 & 15.41 & 50.09 & 17.30 & 49.08 & 16.43 & 30.87 & 13.54 \\
GPT-2 Medium & 25.51 & 22.22 & 48.60 & 16.65 & 49.91 & 18.72 & 49.08 & 18.78 & 30.78 & 14.45 \\
GPT-2 Large & 24.59 & 21.31 & 48.70 & 16.98 & 49.53 & 20.51 & 48.62 & 20.33 & 31.45 & 15.86 \\
GPT-2 XL & 29.03 & 20.99 & 50.10 & 18.00 & 49.81 & 20.53 & 49.20 & 20.39 & 49.09 & 16.98 \\
GPT-Neo 125M & 25.83 & 15.38 & 48.90 & 15.09 & 49.91 & 18.38 & 49.66 & 17.05 & 30.87 & 12.17 \\
GPT-Neo 1.3B & 27.70 & 18.08 & 48.30 & 16.43 & 49.91 & 17.90 & 50.34 & 17.44 & 37.68 & 14.24 \\
GPT-Neo 2.7B & 23.16 & 20.03 & 51.20 & 16.73 & 45.97 & 18.15 & 46.79 & 18.30 & 69.03 & 15.01 \\
GPT-J 6B & 25.03 & 22.72 & 51.20 & 18.33 & 50.47 & 17.40 & 50.57 & 17.32 & 65.29 & 15.58 \\
Phi 1 & 22.74 & 30.26 & 48.30 & 20.73 & 50.38 & 23.37 & 52.64 & 22.33 & 66.92 & 16.14 \\
Phi 1.5 & 59.53 & 22.92 & 70.20 & 18.61 & 50.47 & 16.32 & 50.92 & 15.74 & 66.25 & 15.30 \\
Phi 2 & 78.80 & 21.38 & 82.70 & 18.80 & 78.61 & 16.23 & 83.37 & 15.50 & 66.92 & 14.58 \\
Gemma 2B & 33.08 & 18.93 & 57.20 & 20.17 & 53.75 & 22.13 & 50.00 & 21.38 & 68.65 & 14.54 \\
Llama2 7B & 54.39 & 18.36 & 62.70 & 19.99 & 59.29 & 13.51 & 64.11 & 14.39 & 42.19 & 12.81 \\
Llama2 7B Chat & 64.46 & 19.20 & 61.60 & 21.37 & 53.85 & 15.52 & 52.98 & 15.43 & 69.03 & 15.65 \\
Llama2 13B & 66.34 & 15.10 & 74.30 & 26.29 & 69.61 & 12.56 & 65.37 & 13.87 & 69.32 & 14.82 \\
Llama2 13B Chat & 69.13 & 16.77 & 79.60 & 18.75 & 74.58 & 14.89 & 77.75 & 13.80 & 69.32 & 14.93 \\
Llama3 8B & 62.42 & 19.61 & 78.20 & 22.51 & 58.63 & 18.60 & 63.30 & 17.43 & 67.11 & 16.22 \\
Llama3 8B Instruct & 79.39 & 14.63 & 84.80 & 15.41 & 77.95 & 9.41 & 80.16 & 9.71 & 68.36 & 11.15 \\
Mistral 7B & 81.72 & 20.05 & 80.10 & 27.67 & 66.14 & 19.91 & 68.69 & 18.55 & 69.42 & 19.41 \\
\bottomrule
\end{tabular}
\end{table*}

\begin{table*}
\caption{The table shows Accuracy and corresponding last layer (MLP Out) Intrinsic Dimensions of various models on MMLU datasets}
 \label{tab:accuracy-intrinsic-table-mmlu-datasets}
 \centering
\begin{tabular}{lcccccccc}
\toprule
\textbf{Model} & \multicolumn{2}{c}{\textbf{STEM}} & \multicolumn{2}{c}{\textbf{Humanities}} & \multicolumn{2}{c}{\textbf{Social Sciences}} & \multicolumn{2}{c}{\textbf{Other}} \\
\cmidrule(lr){2-3}\cmidrule(lr){4-5}\cmidrule(lr){6-7}\cmidrule(lr){8-9}
 & \textbf{Acc} & \textbf{ID} & \textbf{Acc} & \textbf{ID} & \textbf{Acc} & \textbf{ID} & \textbf{Acc} & \textbf{ID} \\
\midrule
Random Baseline & 25 & - & 25 & - & 25 & - & 25 & - \\
GPT-2 & 21.37 & 16.53 & 24.25 & 16.69 & 21.81 & 18.55 & 23.60 & 17.80 \\
GPT-2 Medium & 21.64 & 17.62 & 24.21 & 17.43 & 21.74 & 19.78 & 23.81 & 19.31 \\
GPT-2 Large & 22.23 & 19.48 & 24.44 & 18.98 & 22.16 & 21.17 & 24.06 & 21.92 \\
GPT-2 XL & 24.62 & 18.50 & 24.31 & 18.13 & 23.46 & 21.24 & 24.46 & 21.46 \\
GPT-Neo 125M & 21.40 & 14.12 & 24.21 & 15.86 & 21.84 & 16.39 & 23.72 & 15.05 \\
GPT-Neo 1.3B & 27.87 & 14.06 & 24.82 & 14.32 & 26.91 & 15.02 & 25.17 & 14.83 \\
GPT-Neo 2.7B & 27.07 & 15.93 & 24.99 & 15.72 & 24.76 & 18.40 & 26.31 & 17.16 \\
GPT-J 6B & 25.78 & 15.30 & 26.70 & 14.35 & 26.03 & 18.11 & 27.21 & 17.02 \\
Phi 1 & 23.56 & 18.97 & 25.89 & 19.99 & 23.63 & 20.53 & 26.31 & 21.05 \\
Phi 1.5 & 31.84 & 15.16 & 33.62 & 15.75 & 41.57 & 14.97 & 40.38 & 16.16 \\
Phi 2 & 44.60 & 18.04 & 47.57 & 15.56 & 63.02 & 17.75 & 58.70 & 20.18 \\
Gemma 2B & 28.73 & 13.84 & 30.50 & 16.14 & 33.86 & 16.37 & 35.44 & 16.92 \\
Llama2 7B & 31.44 & 13.69 & 36.37 & 14.84 & 42.67 & 15.30 & 43.24 & 15.49 \\
Llama2 7B Chat & 35.65 & 16.27 & 42.64 & 16.54 & 51.48 & 15.39 & 52.34 & 16.78 \\
Llama2 13B & 41.78 & 15.88 & 45.93 & 15.95 & 56.52 & 18.07 & 56.23 & 18.03 \\
Llama2 13B Chat & 40.69 & 15.21 & 46.93 & 14.81 & 60.81 & 15.35 & 59.22 & 15.49 \\
Llama3 8B & 50.07 & 14.83 & 49.22 & 17.65 & 67.31 & 17.21 & 66.13 & 15.70 \\
Llama3 8B Instruct & 50.40 & 12.70 & 49.99 & 13.03 & 69.78 & 13.63 & 67.80 & 13.09 \\
Mistral 7B & 48.05 & 18.22 & 52.22 & 15.91 & 67.86 & 22.86 & 66.78 & 22.25 \\
\bottomrule
\end{tabular}
\end{table*}

\begin{table}
  \caption[Correlation Table for LLama Models]{The table highlights the correlation between ID estimates of the last layer (MLP Out) and the corresponding model performance for 
LLama models. The negative correlation indicates ID being lower for better-performing models, making ID estimates a weak proxy for the model's generalization capabilities.}
  \label{tab:llama-correlation-results}
  \centering
  \begin{tabular}{lcccc}
    \toprule
    \textbf{Dataset} & \textbf{Pearson} & \textbf{Kendall Tau}  & \textbf{Spearman} & \textbf{Accuracy}  \\
    \midrule
COPA & $-0.356$ & $-0.467$ & $-0.543$ & $73.53 $ \\
COLD & $-0.202$ & $-0.333$ & $-0.486$ & $69.16$ \\
Rotten Tomatoes & $-0.673$ & $-0.600$ & $-0.771$ & $65.65$ \\
SST2 & $-0.720$ & $-0.867$ & $-0.943$ & $67.27$ \\
\midrule
AG News & $-0.709$ & $-0.600$ & $-0.771$ & $66.02$ \\ 
MMLU STEM & $-0.309$ & $-0.333$ & $-0.371$ & $41.67$ \\
MMLU Humanities & $-0.022$ & $-0.200$ & $-0.257$ & $45.18$ \\
MMLU Social Sciences & $-0.084$ & $-0.067$ & $-0.143$ & $58.09$ \\
MMLU Other & $-0.438$ & $-0.276$ & $-0.406$ & $57.50$ \\

    \bottomrule
  \end{tabular}
\end{table}

\begin{table}
    \centering
    \caption[Correlation Table for well-performing Models]{Correlation between ID estimates of the last layer (MLP Out) and the corresponding model performances for all the models which perform better than the baseline, i.e., 25\% or 50\% for 4 options and 2 options datasets respectively.}
    \label{tab:correlation-all-models}
    \begin{tabular}{lccccc}
    \toprule
    \textbf{Dataset} & \textbf{Pearson} & \textbf{Kendall Tau}& \textbf{Spearman}& \textbf{Accuracy}\\
    \midrule
COPA & $-0.02$ & $-0.14$ & $-0.18$ & $74.47$ \\ 
COLD & $-0.75$ & $-0.60$ & $-0.71$ & $73.90$ \\ 
Rotten Tomatoes & $-0.81$ & $-0.67$ & $-0.80$ & $72.07$ \\ 
SST2 & $-0.66$ & $-0.60$ & $-0.66$ & $69.90$ \\ 
\midrule
AG News & $-0.28$ & $-0.14$ & $-0.12$ & $63.87$ \\ 
MMLU STEM & $0.04$ & $-0.14$ & $-0.21$ & $42.58$ \\ 
MMLU Humanities & $-0.09$ & $-0.21$ & $-0.29$ & $44.22$ \\ 
MMLU Social Sciences & $0.25$ & $0.07$ & $0.05$ & $56.28$ \\ 
MMLU Other & $0.06$ & $-0.11$ & $-0.18$ & $55.90$ \\ 

    \bottomrule
    \end{tabular}
\end{table}

\begin{table}
    \centering
    \caption[Figure Reference Table]{The table provides reference links to the figures corresponding to the correlation between the trend of intrinsic dimension across the relative depth of the model and the relationship between accuracy and intrinsic dimension along the depth of the model.}
    \label{app-tab:plots-ref-table}
    \begin{tabular}{lcc}
    \toprule
    \textbf{Dataset} & \textbf{Corr. Matrix (Ref.)} & \textbf{Accuracy-ID (Ref.)}\\
    \midrule
AGNews \cite{zhang2016characterlevelconvolutionalnetworkstext} & Figure \ref{app-fig:interpolated-correlation-agnews} & Figure \ref{fig:peak-trend-AGNews}\\ 
CoLA \cite{warstadt2019neural}                             & Figure  \ref{app-fig:interpolated-correlation-cola}          & -  \\ 
COLD \cite{joshi2024cold}                              & Figure \ref{app-fig:interpolated-correlation-cold}         & Figure \ref{fig:peak-trend-COLD}  \\
RottenTomatoes \cite{rottentomatoes}                    & Figure  \ref{app-fig:interpolated-correlation-rottentomatoes} & Figure \ref{fig:peak-trend-RottenTomatoes} \\ 
SST-2  \cite{socher-etal-2013-recursive}                            & 
Figure  \ref{app-fig:interpolated-correlation-sst2}   & Figure \ref{fig:peak-trend-SST2}         \\ 
MMLU STEM \cite{hendrycks2021measuringmassivemultitasklanguage}                              & Figure  \ref{app-fig:interpolated-correlation-mmlu-stem}  & Figure \ref{fig:peak-trend-MMLUSTEM}          \\ 
MMLU Humanities \cite{hendrycks2021measuringmassivemultitasklanguage}                              & Figure  \ref{app-fig:interpolated-correlation-mmlu-humanities}    &  Figure \ref{fig:peak-trend-MMLU-Humanities}        \\ 
MMLU Social Sciences \cite{hendrycks2021measuringmassivemultitasklanguage}                              & Figure  \ref{app-fig:interpolated-correlation-mmlu-social-sciences}   &  Figure \ref{fig:peak-trend-MMLU-Social}         \\ 
MMLU Other \cite{hendrycks2021measuringmassivemultitasklanguage}                              & Figure  \ref{app-fig:interpolated-correlation-mmlu-other}   & Figure \ref{fig:peak-trend-MMLU-Other}        \\ 
COPA \cite{gordon-etal-2012-semeval}                             & Figure  \ref{app-fig:interpolated-correlation-copa}  & Figure \ref{fig:peak-trend-COPA}\\
\bottomrule
    \end{tabular}
\end{table}

\subsection{Compute Resources} \label{app-sec:compute-resources}
We perform all the experiments using a machine with 1 NVIDIA A40 GPU. We use only the
open-weight models with frozen parameters to present the results for better reproducibility in the future.

\subsection{Additional Results and Discussion}
\begin{sidewaysfigure}
  \centering
  {
  \includegraphics[scale=0.37]{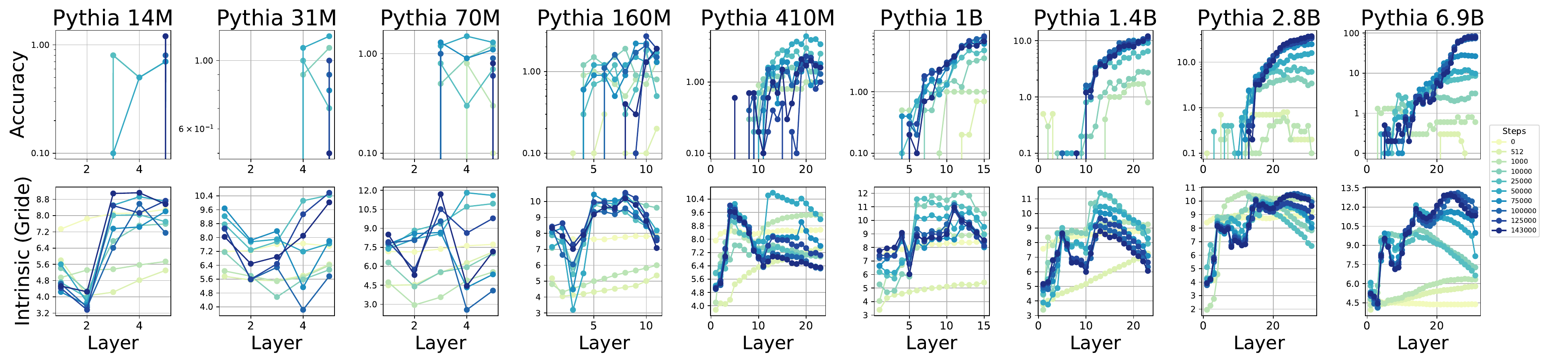}}
 \caption[Intrinsic dimension and accuracy trend of Pythia models on Arithmetic Dataset]{
 The figure shows ID estimates for residual post layers in Pythia series models as the training processes for the Arithmetic Dataset. We observe all the models with different sizes following a similar trend of manifold evolution as the training progresses, finally converging to a similar characteristic hunchback-like shape.   
 }
 \label{app-fig:arithmetic-peak-trend}
\end{sidewaysfigure}

\begin{sidewaysfigure}
  \centering
   \includegraphics[scale=0.4]{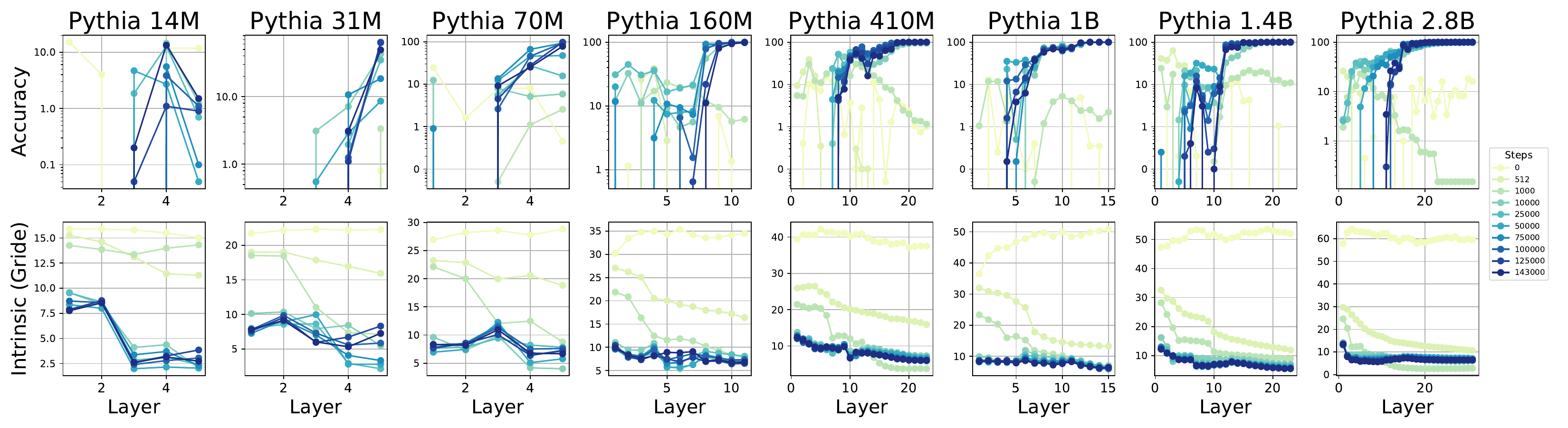}
 \caption[Intrinsic dimension and accuracy trend of Pythia models on Greater Than Dataset]{
  The figure shows ID estimates for residual post layers in Pythia series models as the training processes for the Greater Than Dataset. We observe all the models with different sizes following a similar trend of manifold evolution as the training progresses. The Greater Than dataset, being straightforward, shows the accuracy boost even from small-scale models, where smaller models show a hunchback-like shape, with minimal changes observed for larger models that show an accuracy boost from initial layers. 
 }
 \label{app-fig:greater-than-peak-trend}
\end{sidewaysfigure}

The intrinsic dimension often relies on a smaller range of (5 - 38) when compared to the extrinsic dimensions (aka model hidden dimensions (768 - 4096)), irrespective of size and number of layers present in these models. We found this trend to be consistent for both template-based synthetic datasets as well as real-world datasets, pointing toward the language-specific tasks being present in low-dimensional manifolds. This suggests that all these networks learn to compress the language information to a lower-dimensional space of a similar range. 
Table \ref{tab:accuracy-intrinsic-table-datasets} and Table \ref{tab:accuracy-intrinsic-table-mmlu-datasets} summarize the Intrinsic dimension (of the last layer, MLP Out) estimates obtained for various datasets and compare them with the performance. 
We highlight the additional trends observed in our experiments below:



\textbf{Manifolds evolving during Training:}
In our setup, we consider two template-based synthetic datasets, the arithmetic and the greater than dataset. The arithmetic dataset was also used for monitoring learning during the training of Pythia series models \cite{biderman2023pythia}. 
Figure \ref{app-fig:arithmetic-peak-trend} shows ID estimates for layers in Pythia series models as the training processes for the Arithmetic Dataset. We observe all the models with different sizes
following a similar trend of manifold evolution as the training progresses, finally converging to a similar characteristic, hunchback-like shape.
On the Greater than dataset, we observe a similar evolution trend. However, the Greater Than dataset, being straightforward (less complex), shows the accuracy boost even from small-scale models. In Figure \ref{app-fig:greater-than-peak-trend}, smaller models show a hunchback-like shape with minimal changes observed for larger models that show an accuracy boost from initial layers.

\begin{figure*}
  \centering
   \includegraphics[width=0.9\textwidth]{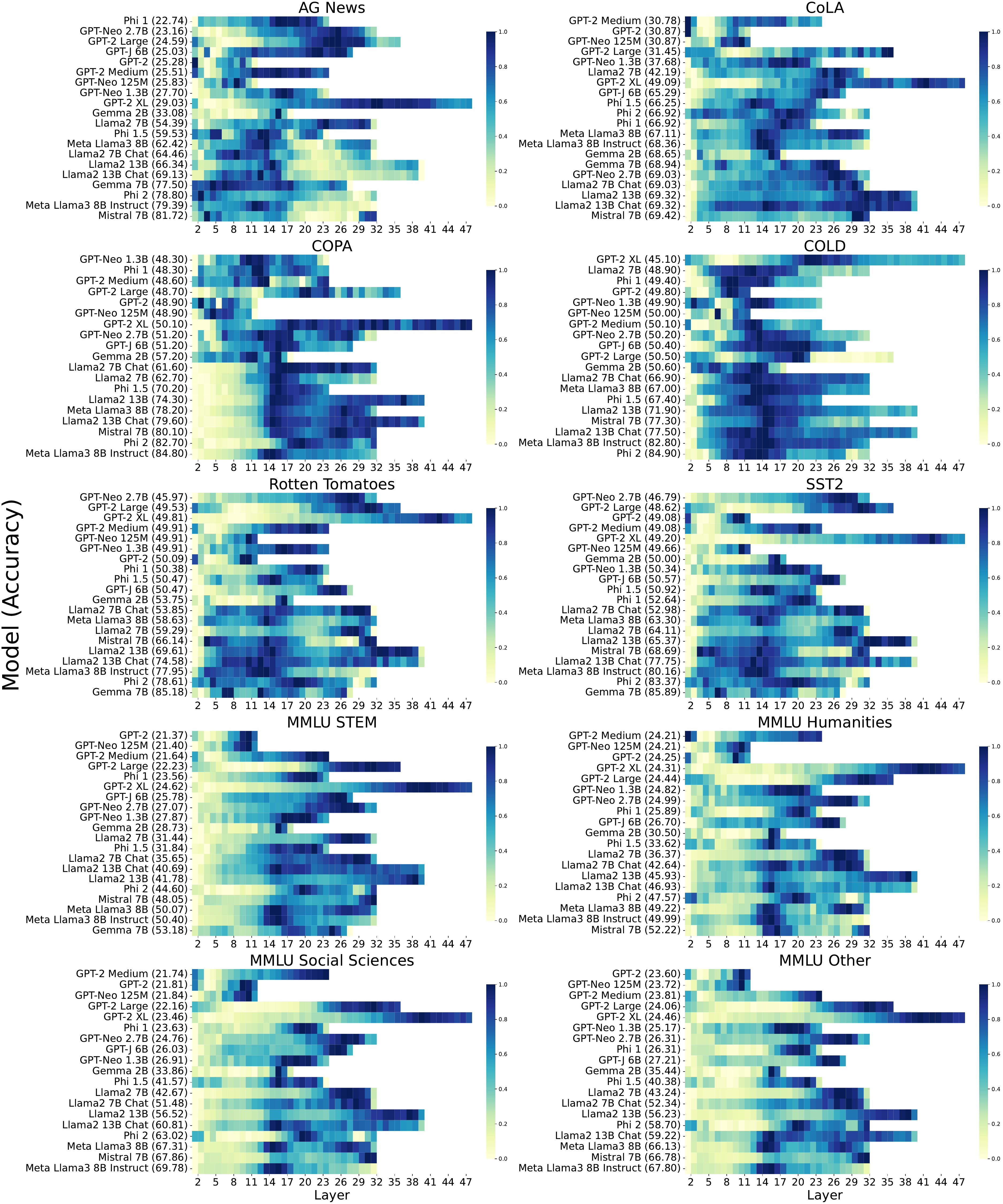}
 \caption[Intrinsic trend across the model (MLP Out) using MLE Estimator]{
  The figure shows a characteristic trend observed for a wide range of models for different real-world datasets. We observe a peak in the middle layers, highlighting a hunchback-like shape. The intrinsic dimensions are estimated using the \textbf{MLE} estimator. The y-label shows the model names along with the accuracy in brackets, sorted from low-performing models to high-performing models from top to bottom.
 }
 \label{app-fig:hunchback-MLE}
\end{figure*}

\begin{figure*}
  \centering
   \includegraphics[width=0.9\textwidth]{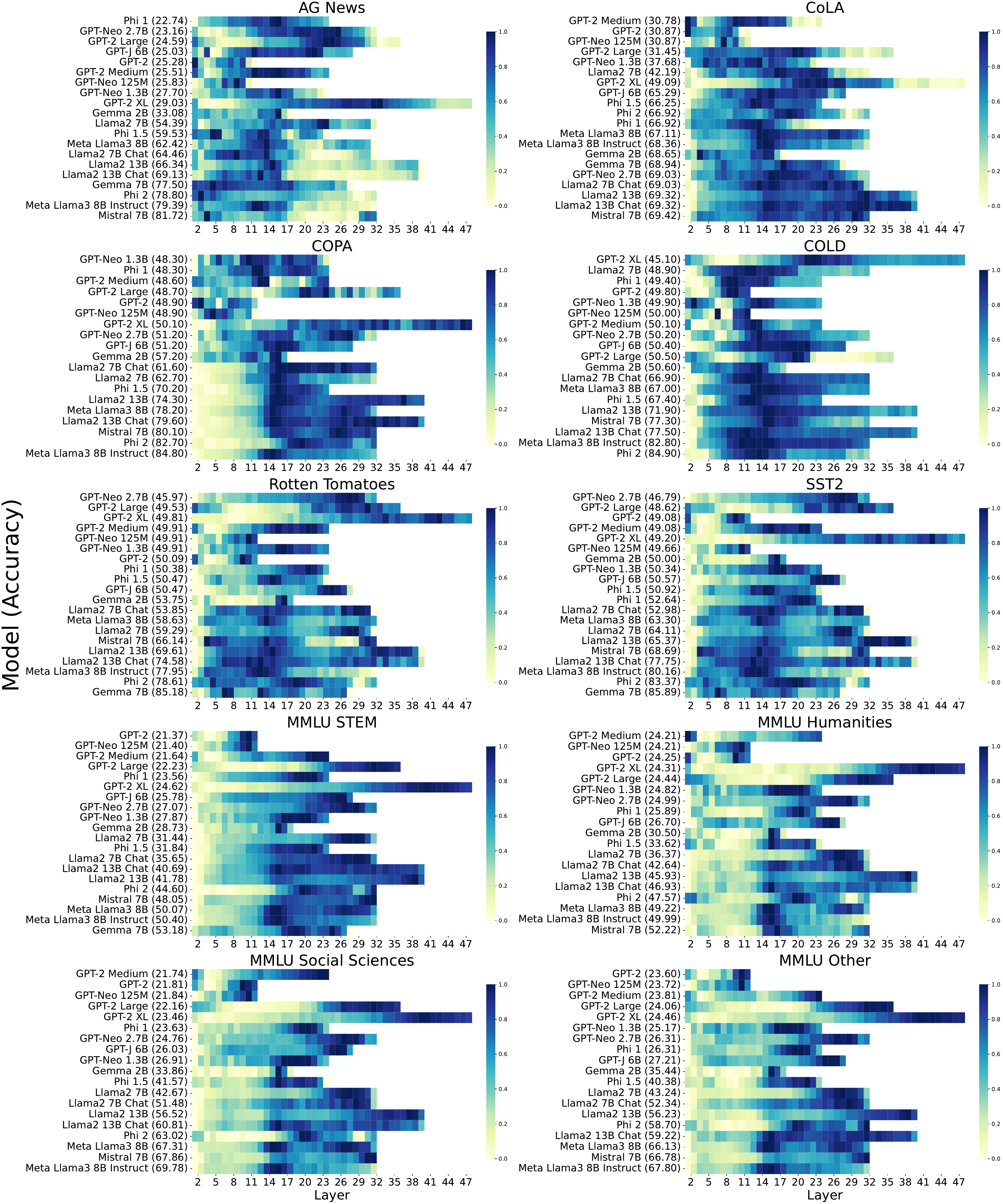}
 \caption[Intrinsic trend across the model (MLP Out) using MLE (harmonic mean) Estimator]{
 The figure shows a characteristic trend observed for a wide range of models for different real-world datasets. We observe a peak in the middle layers, highlighting a hunchback-like shape. The intrinsic dimensions are estimated using the modified version of the \textbf{MLE (harmonic mean)} estimator. The y-label shows the model names along with the accuracy in brackets, sorted from low-performing models to high-performing models from top to bottom.
 }
 \label{app-fig:hunchback-MLE-Modified}
\end{figure*}

\begin{figure*}
  \centering
   \includegraphics[width=0.9\textwidth]{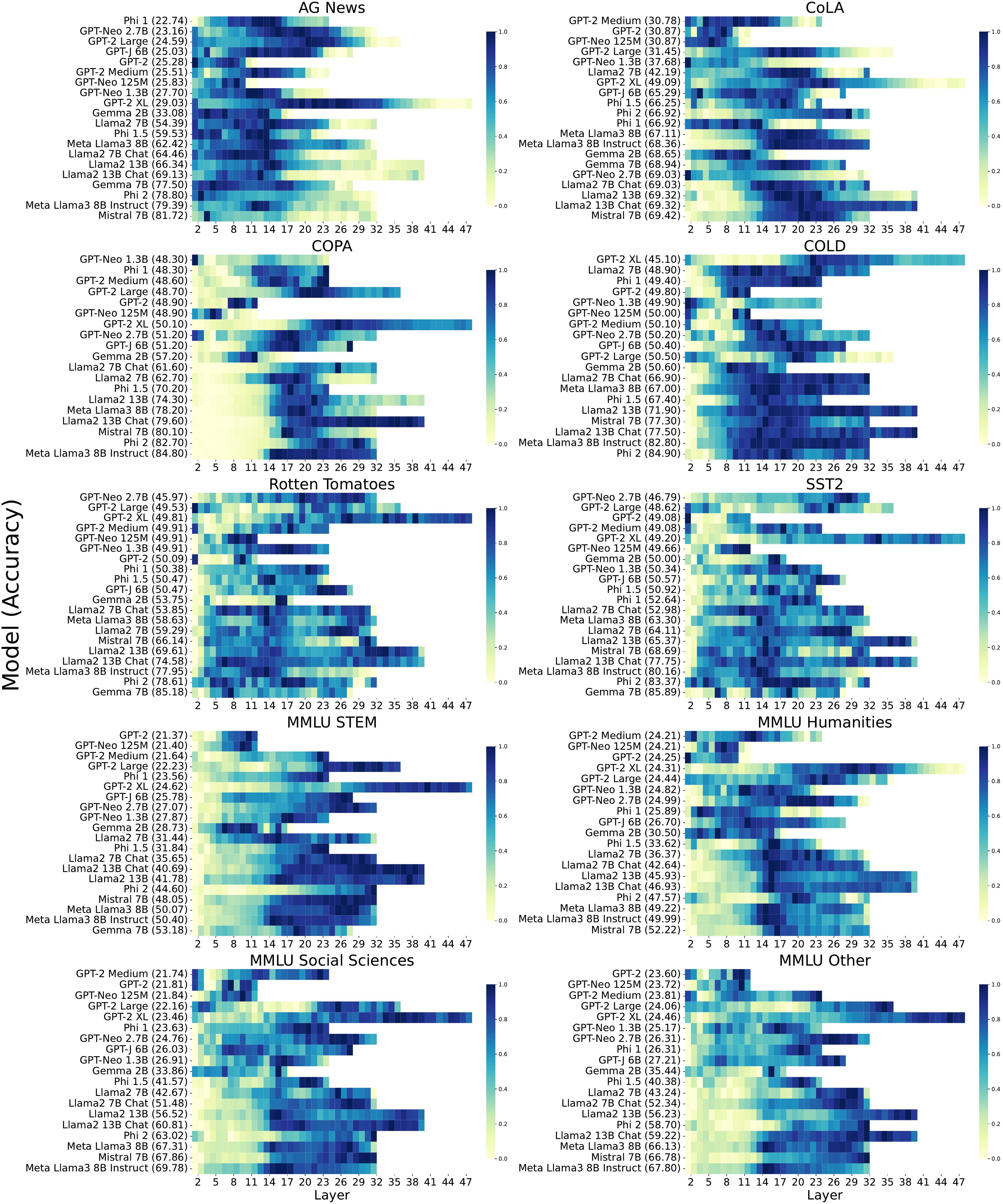}
 \caption[Intrinsic trend across the model (MLP Out) using TwoNN Estimator]{The figure shows a characteristic trend observed for a wide range of models for different real-world datasets. We observe a peak in the middle layers, highlighting a hunchback-like shape. The intrinsic dimensions are estimated using the \textbf{TwoNN} estimator. The y-label shows the model names along with the accuracy in brackets, sorted from low-performing models to high-performing models from top to bottom.}
 \label{app-fig:hunchback-TwoNN}
\end{figure*}

\begin{figure*}
  \centering
   \includegraphics[width=0.9\textwidth]{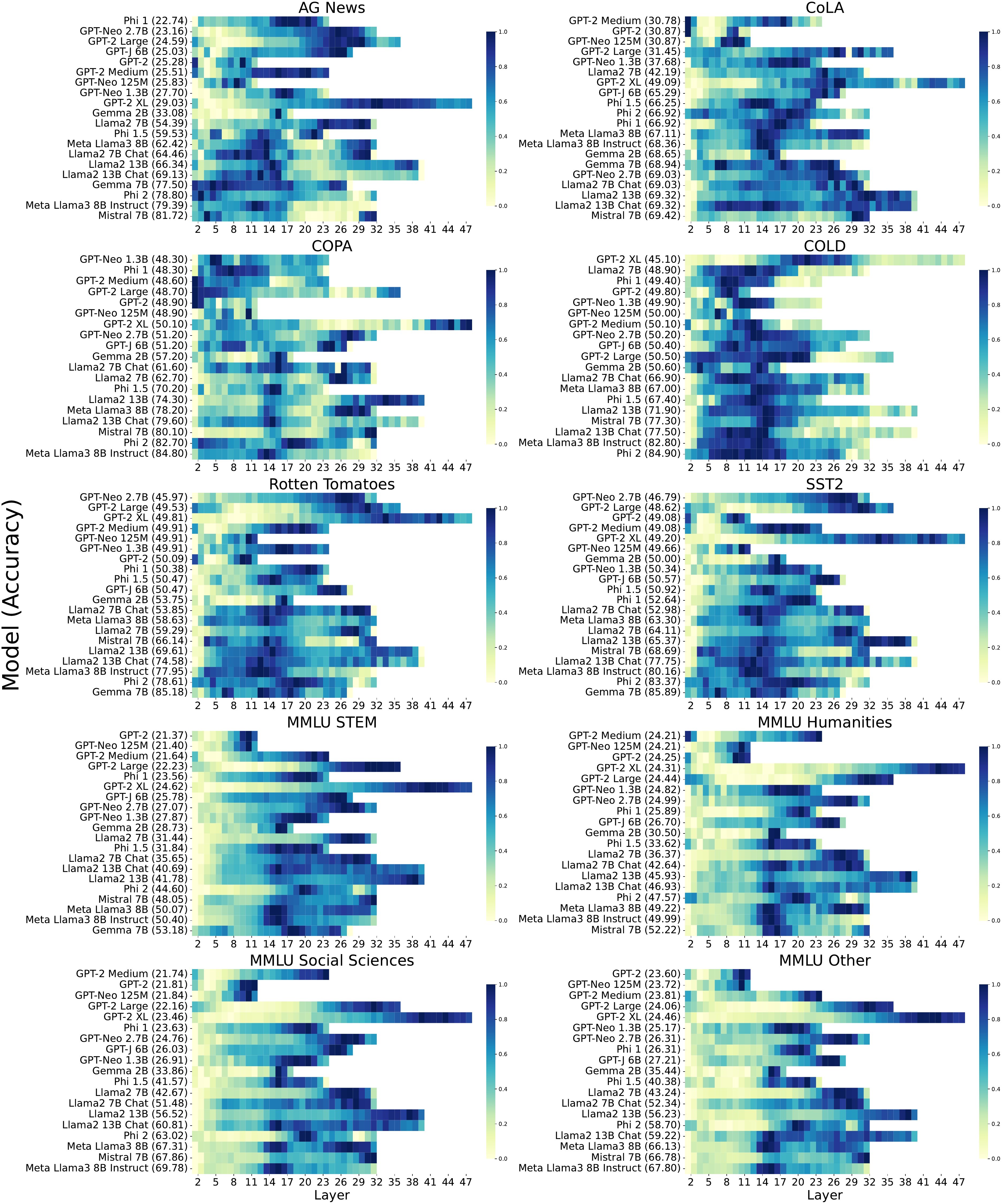}
 \caption[Intrinsic trend across the model (MLP Out) using GRIDE Estimator]{The figure shows a characteristic trend observed for a wide range of models for different real-world datasets. We observe a peak in the middle layers, highlighting a hunchback-like shape. The intrinsic dimensions are estimated using the \textbf{GRIDE} estimator. The y-label shows the model names along with the accuracy in brackets, sorted from low-performing models to high-performing models from top to bottom.}
 \label{app-fig:hunchback-Gride}
\end{figure*}

\textbf{Characteristic Trend in Latent Layers:} 
In all our experiments for real-world datasets, we observe a characteristic hunchback-like trend in ID estimates across the model's MLP Out layers.
The Figures \ref{app-fig:hunchback-MLE}, \ref{app-fig:hunchback-MLE-Modified}, \ref{app-fig:hunchback-TwoNN}, and \ref{app-fig:hunchback-Gride} show the trends for MLE, MLE modified, TwoNN, and GRIDE estimates, respectively.  
All the estimators show a characteristic trend observed for a wide range of models for different real-world datasets. We observe a
peak in the middle layers, highlighting a hunchback-like shape. 
The y-label shows the model names along with the accuracy in brackets, sorted from low-performing models to high-performing models from
top to bottom. In general, we observe a similar trend being followed for similar performing models. Additionally, the relationship between accuracy and intrinsic dimension (both Residual post and MLP out) is captured by the Figures referred to in the App. \ref{app-tab:plots-ref-table}

\begin{figure*}
  \centering
   \includegraphics[width=0.66\textwidth]{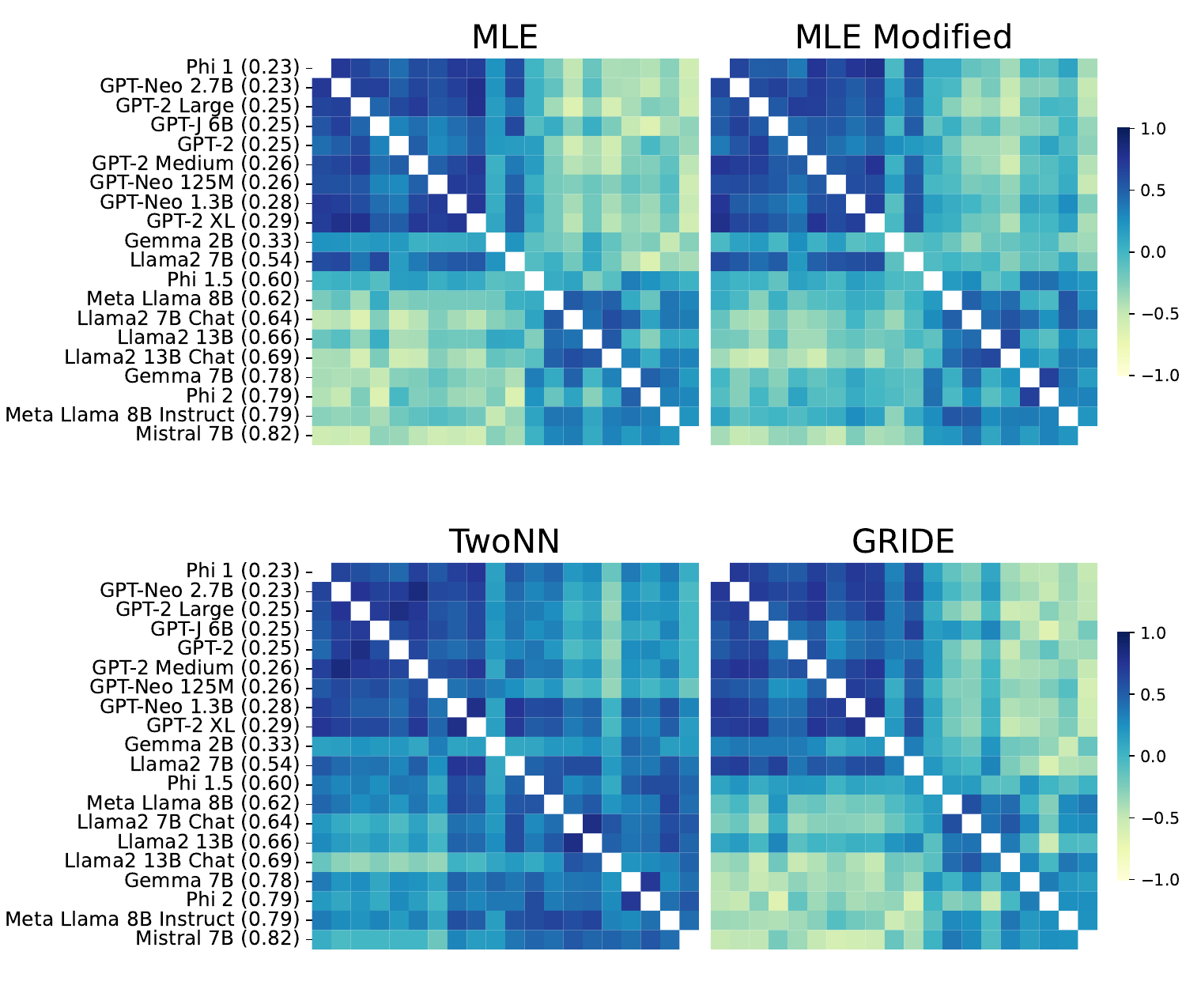}
 \caption[Correlation matrix AG News dataset]{
 The figure shows a correlation between the intrinsic dimensions trajectories throughout layers of multiple open-weight models, denoting the correlation of IDs for all four ID estimators for 
 \textbf{AG News}
 dataset. 
 }
 \label{app-fig:interpolated-correlation-agnews}
\end{figure*}

\begin{figure*}
  \centering
   \includegraphics[width=0.66\textwidth]{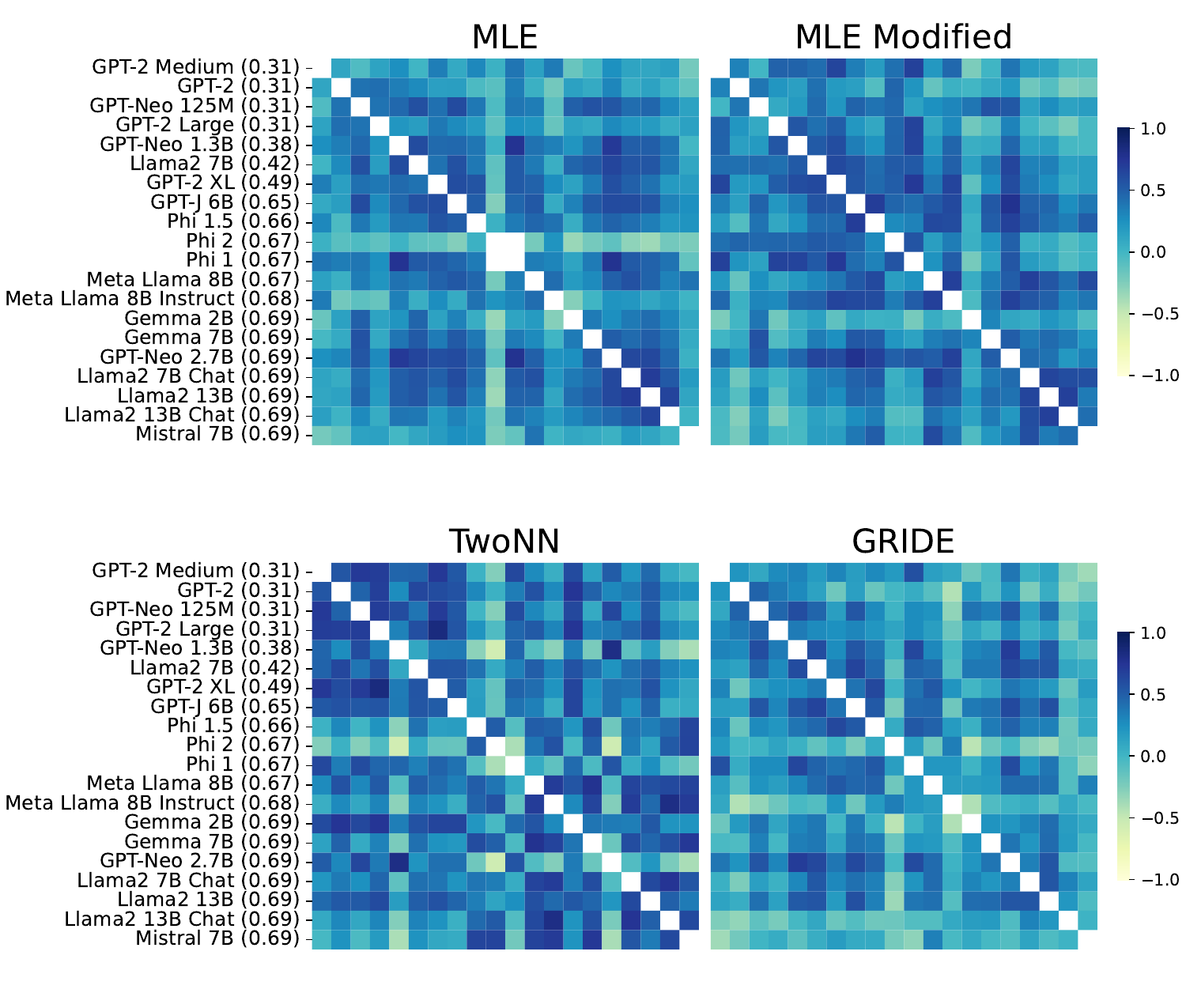}
 \caption[Correlation matrix CoLA dataset]{
 The figure shows a correlation between the intrinsic dimensions trajectories throughout layers of multiple open-weight models, denoting the correlation of IDs for all four ID estimators for 
 \textbf{CoLa}
 dataset. 
 }
 \label{app-fig:interpolated-correlation-cola}
\end{figure*}

\begin{figure*}
  \centering
   \includegraphics[width=0.66\textwidth]{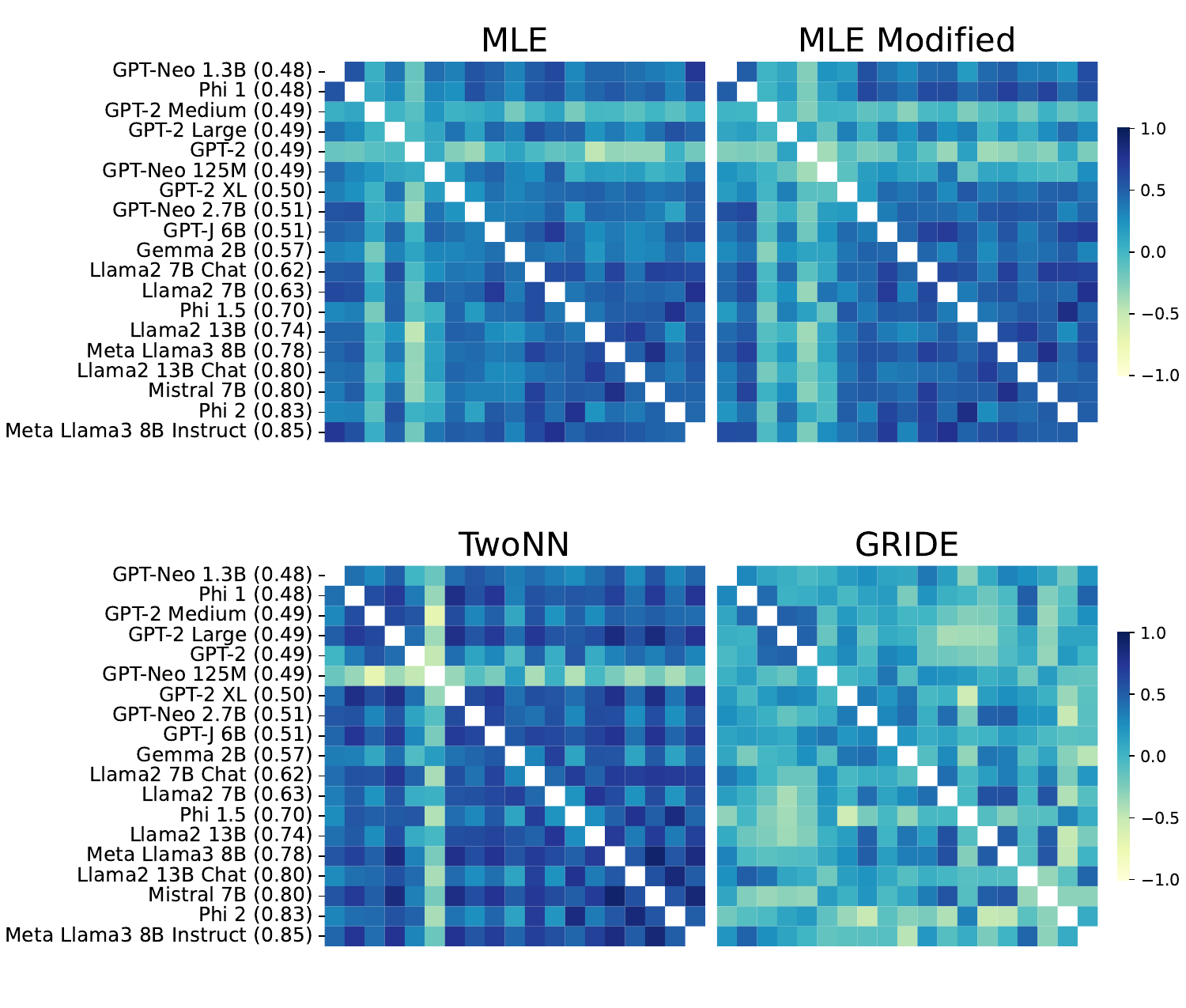}
 \caption[Correlation matrix COPA dataset]{The figure shows a correlation between the intrinsic dimensions trajectories throughout layers of multiple open-weight models, denoting the correlation of IDs for all four ID estimators for 
 \textbf{COPA}
 dataset. 
 }
 \label{app-fig:interpolated-correlation-copa}
\end{figure*}

\begin{figure*}
  \centering
   \includegraphics[width=0.66\textwidth]{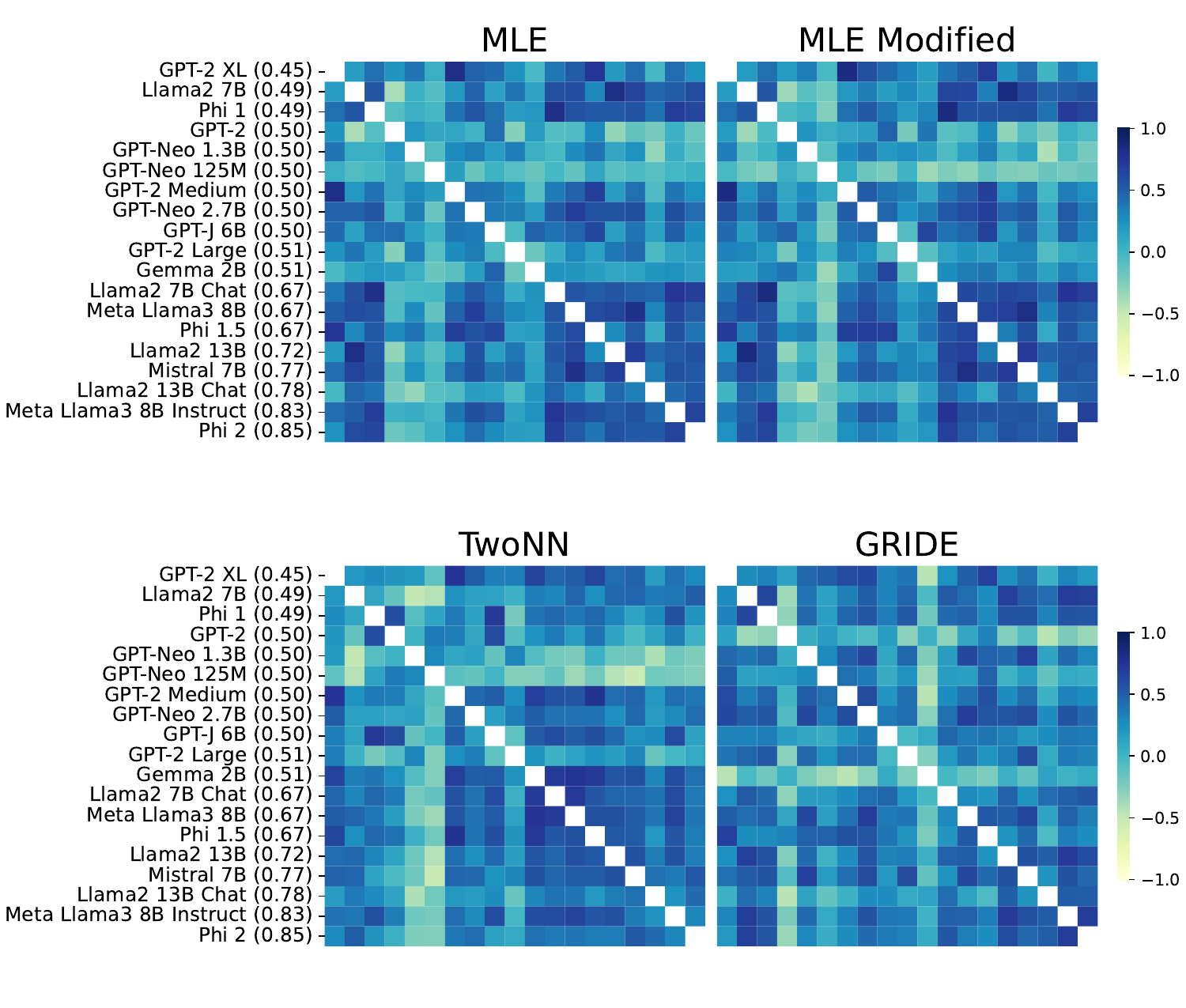}
 \caption[Correlation matrix COLD dataset]{The figure shows a correlation between the intrinsic dimensions trajectories throughout layers of multiple open-weight models, denoting the correlation of IDs for all four ID estimators for 
 \textbf{COLD}
 dataset. 
 }
 \label{app-fig:interpolated-correlation-cold}
\end{figure*}

\begin{figure*}
  \centering
   \includegraphics[width=0.66\textwidth]{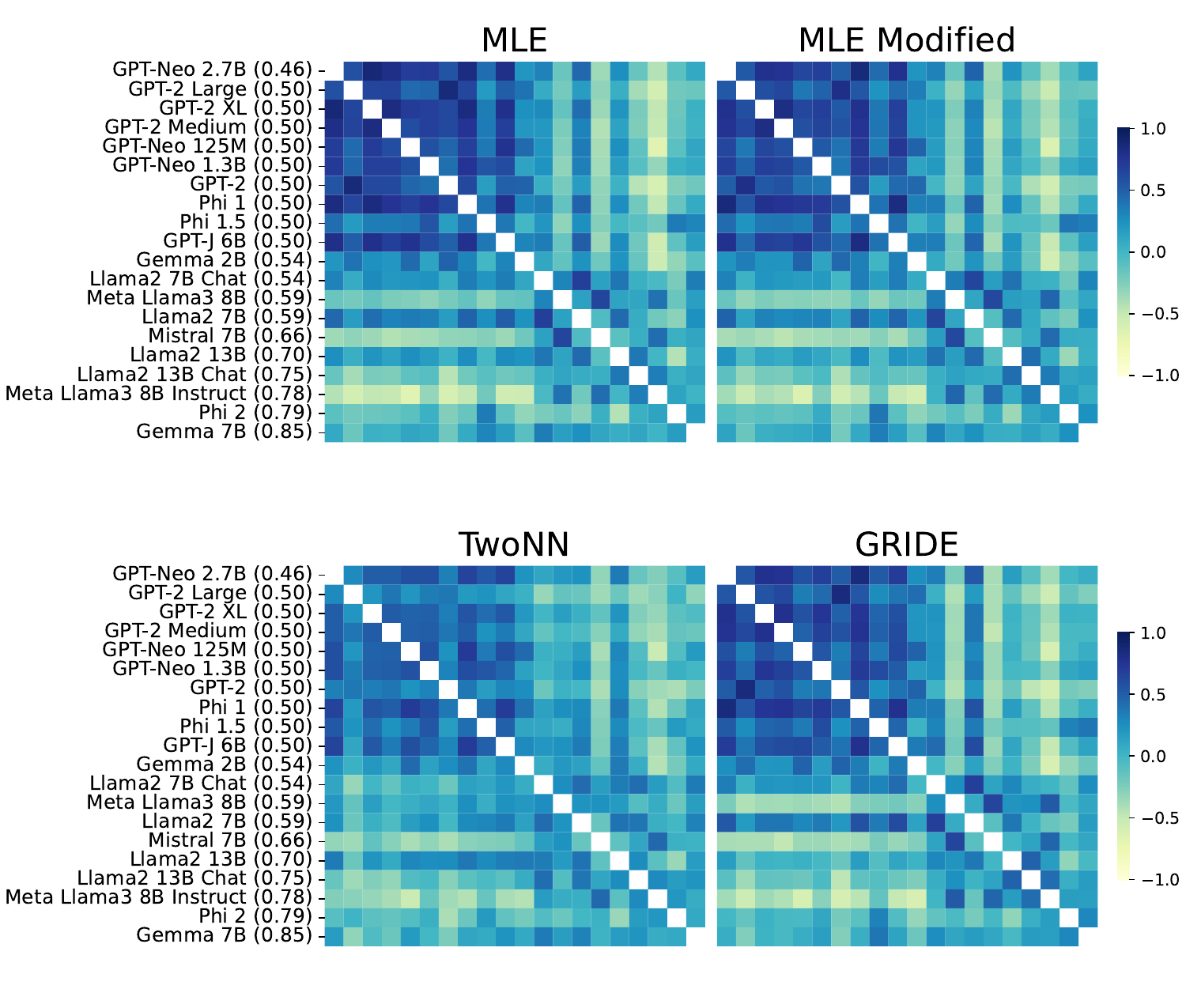}
 \caption[Correlation matrix Rotten Tomatoes dataset]{The figure shows a correlation between the intrinsic dimensions trajectories throughout layers of multiple open-weight models, denoting the correlation of IDs for all four ID estimators for 
 \textbf{Rotten Tomatoes}
 dataset. 
 }
 \label{app-fig:interpolated-correlation-rottentomatoes}
\end{figure*}

\begin{figure*}
  \centering
   \includegraphics[width=0.66\textwidth]{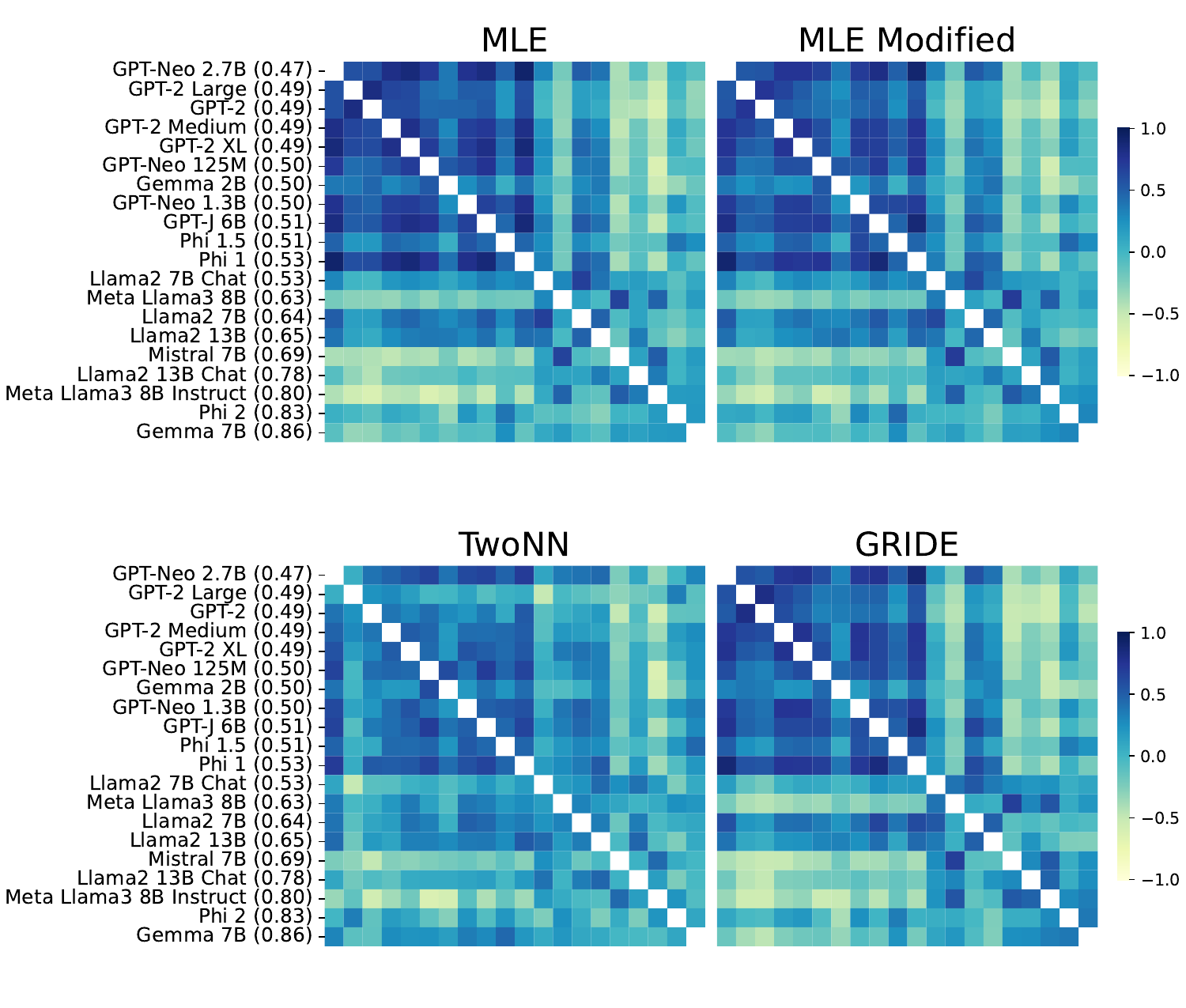}
 \caption[Correlation matrix SST2 dataset]{The figure shows a correlation between the intrinsic dimensions trajectories throughout layers of multiple open-weight models, denoting the correlation of IDs for all four ID estimators for 
 \textbf{SST2}
 dataset. 
 }
 \label{app-fig:interpolated-correlation-sst2}
\end{figure*}

\begin{figure*}
  \centering
   \includegraphics[width=0.66\textwidth]{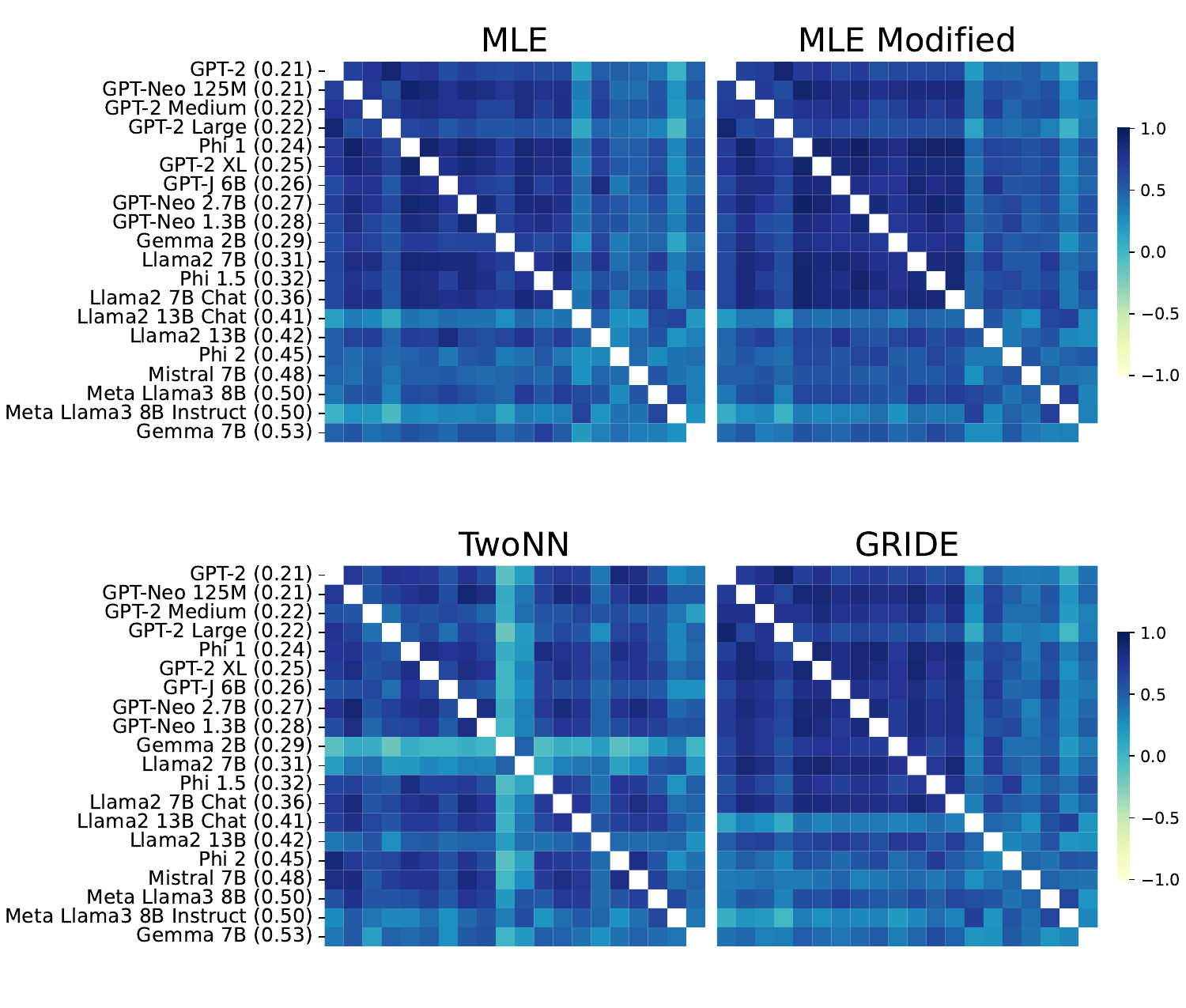}
 \caption[Correlation matrix MMLU STEM dataset]{The figure shows a correlation between the intrinsic dimensions trajectories throughout layers of multiple open-weight models, denoting the correlation of IDs for all four ID estimators for 
 \textbf{MMLU STEM}
 dataset. 
 }
 \label{app-fig:interpolated-correlation-mmlu-stem}
\end{figure*}

\begin{figure*}
  \centering
   \includegraphics[width=0.66\textwidth]{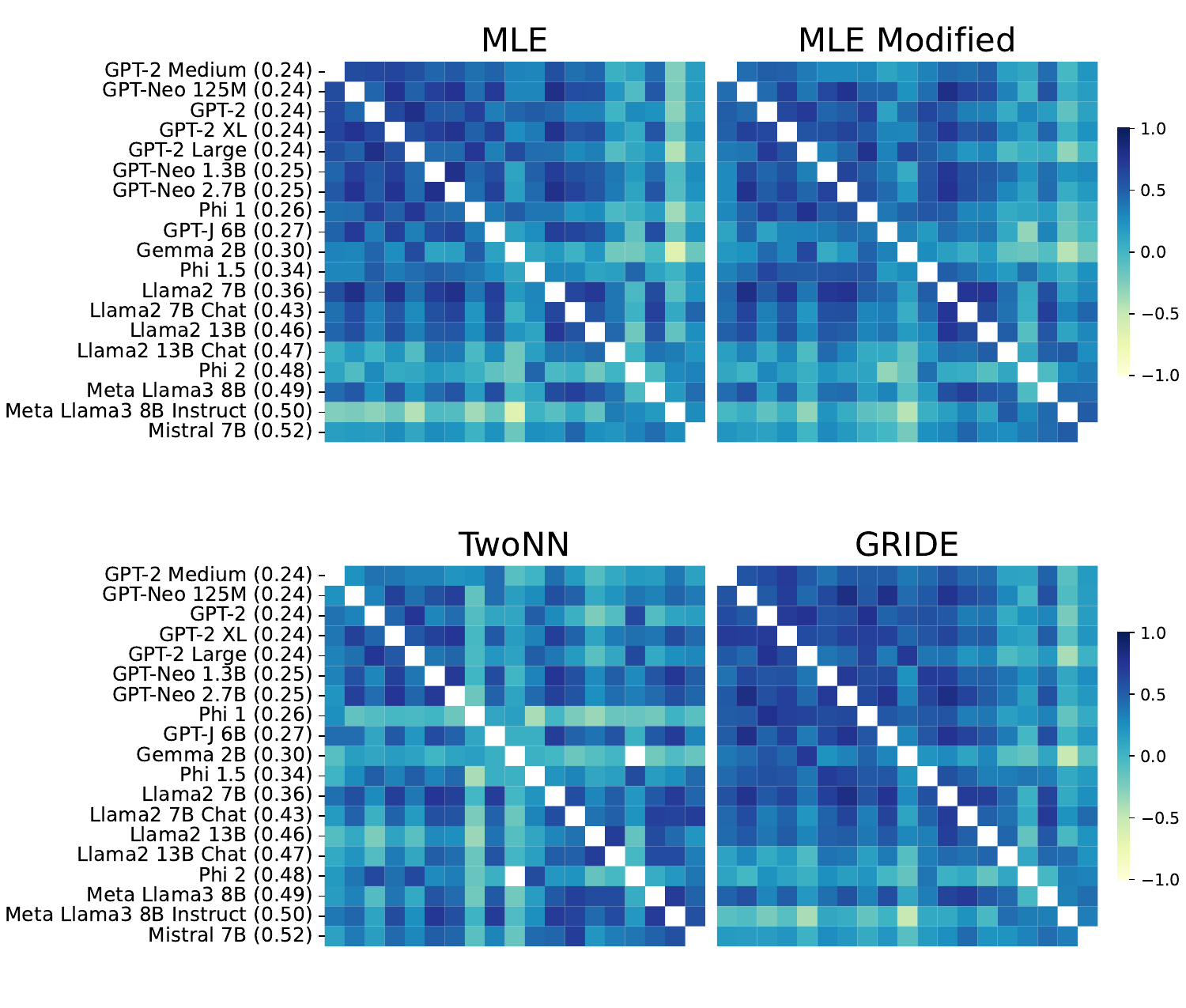}
 \caption[Correlation matrix MMLU Humanities dataset]{The figure shows a correlation between the intrinsic dimensions trajectories throughout layers of multiple open-weight models, denoting the correlation of IDs for all four ID estimators for 
 \textbf{MMLU Humanities}
 dataset. 
 }
 \label{app-fig:interpolated-correlation-mmlu-humanities}
\end{figure*}

\begin{figure*}
  \centering
   \includegraphics[width=0.66\textwidth]{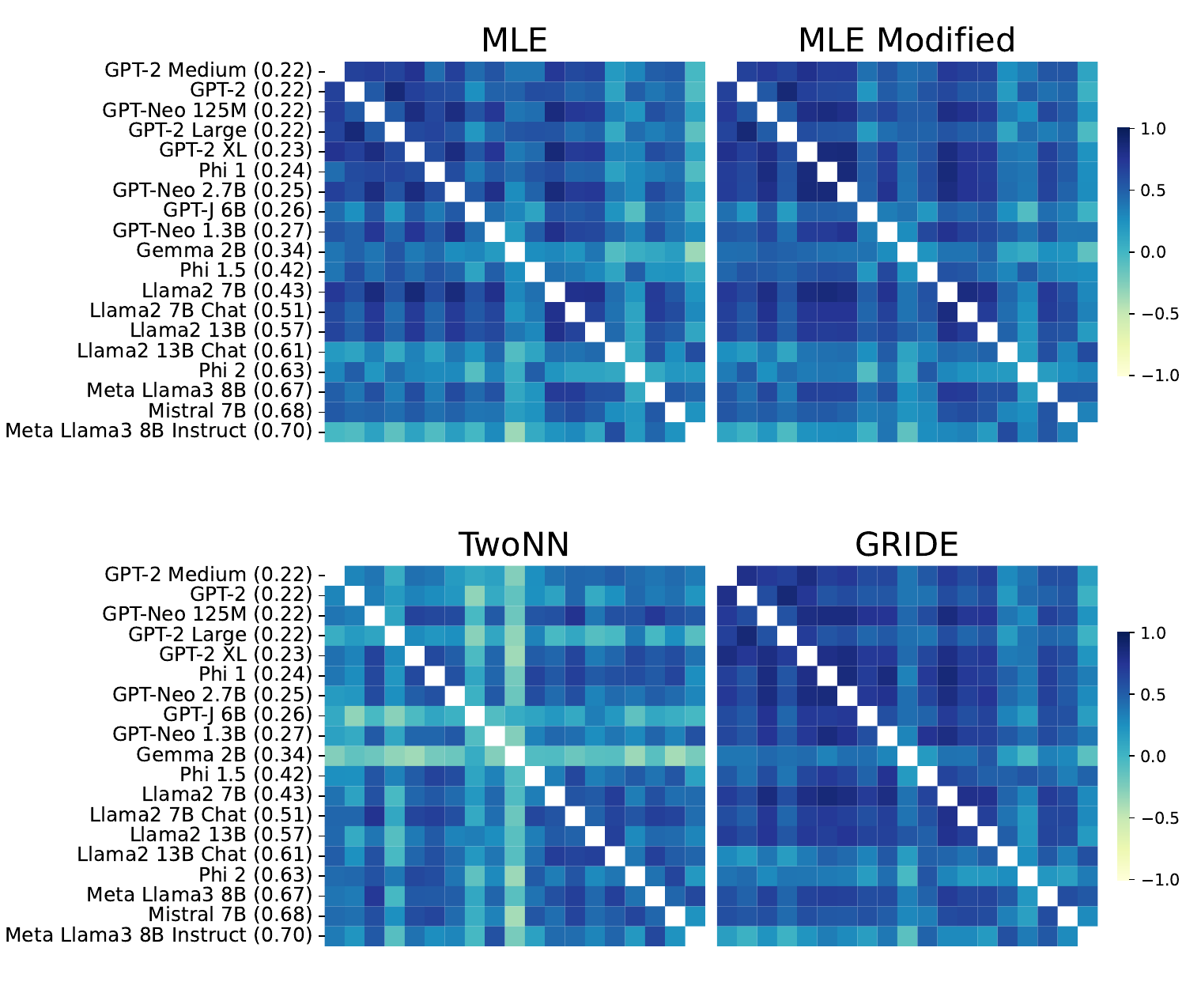}
 \caption[Correlation matrix MMLU Social Sciences dataset]{
 The figure shows a correlation between the intrinsic dimensions trajectories throughout layers of multiple open-weight models, denoting the correlation of IDs for all four ID estimators for 
 \textbf{MMLU Social Sciences}
 dataset. 
 }
 \label{app-fig:interpolated-correlation-mmlu-social-sciences}
\end{figure*}

\begin{figure*}
  \centering
   \includegraphics[width=0.66\textwidth]{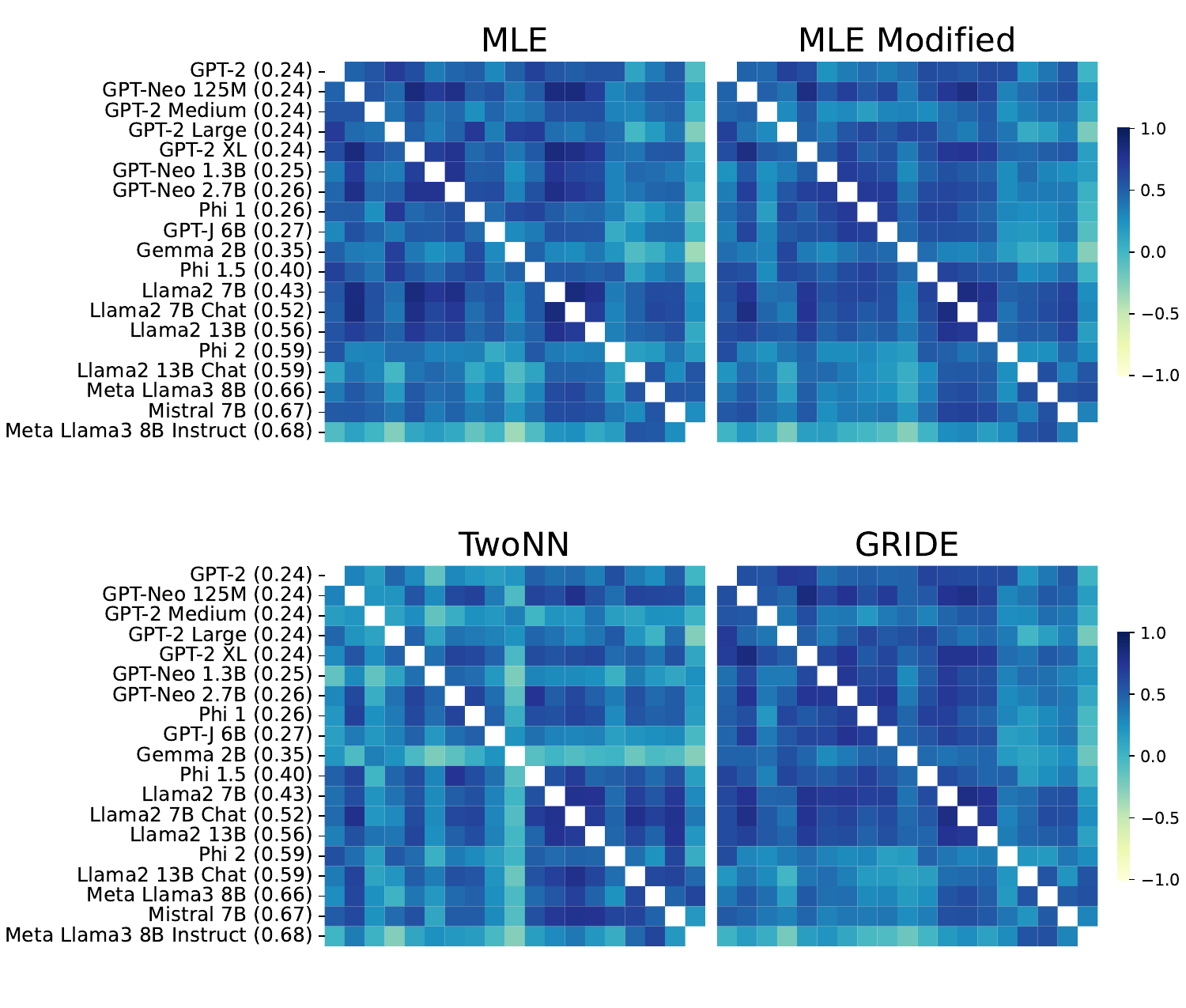}
 \caption[Correlation matrix MMLU Other dataset]{ The figure shows a correlation between the intrinsic dimensions trajectories throughout layers of multiple open-weight models, denoting the correlation of IDs for all four ID estimators for 
 \textbf{MMLU Other}
 dataset. }
 \label{app-fig:interpolated-correlation-mmlu-other}
\end{figure*}

\textbf{Correlation between Characteristic trend in different datasets:}
Figures \ref{app-fig:interpolated-correlation-agnews} to \ref{app-fig:interpolated-correlation-mmlu-other} show a correlation between the intrinsic dimensions estimated for the hidden layers of multiple open-weight models corresponding to MLP Out. 
As the number of layers varies in different models, the estimated ID values were interpolated, considering the notion of relative depth in models. The figures show the correlation of IDs for all four ID estimators. (See table \ref{app-tab:plots-ref-table} for easier reference to figures for different datasets.
Overall, we find a high correlation for similar performing models, highlighting similar trends of IDs observed for multiple models.

\begin{figure*}
\begin{center}
\centerline{\includegraphics[width=\textwidth]{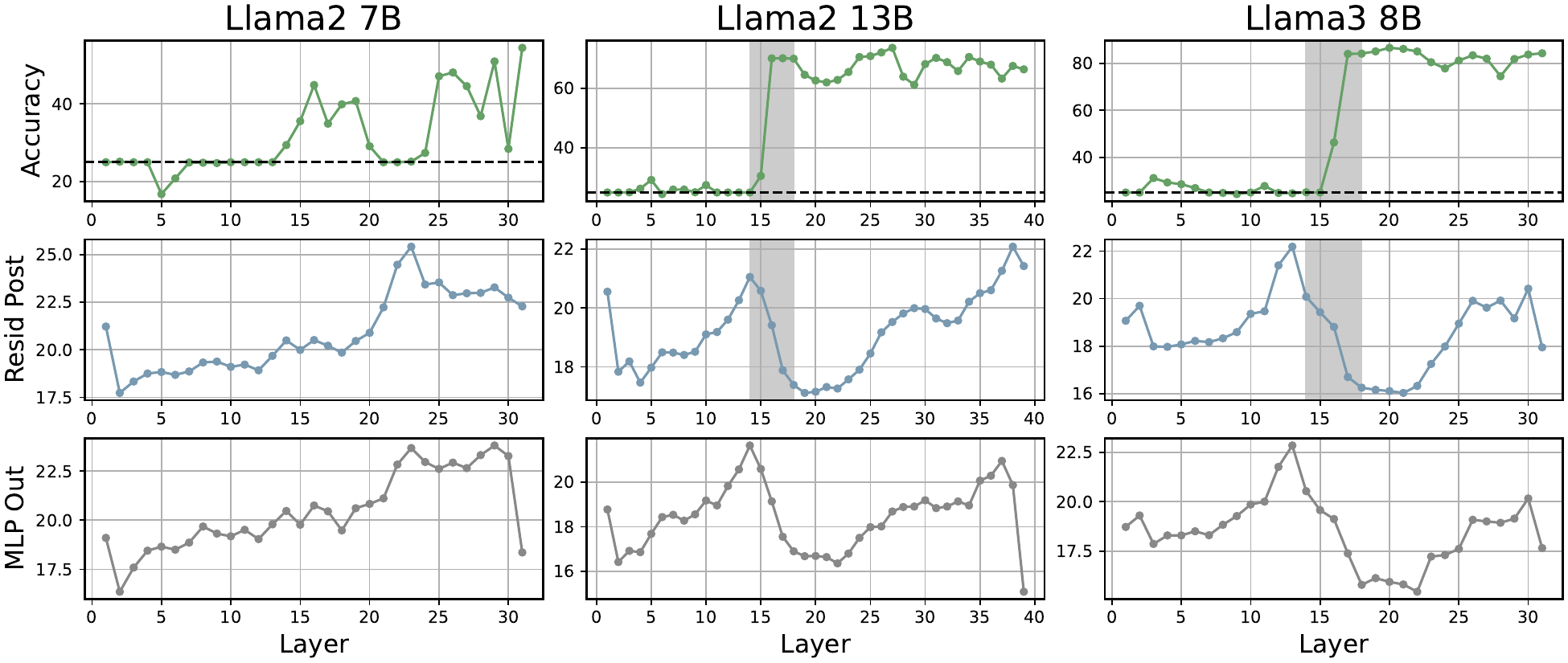}}
\vskip -0.1in
\caption[Accuracy and Intrinsic trends in AGNews]{The figure shows accuracy along with ID trends for different variants of the LLama model on the \textbf{AGNews} dataset}
\label{fig:peak-trend-AGNews}
\end{center}
\vskip -0.35in
\end{figure*}

\begin{figure*}
\begin{center}
\centerline{\includegraphics[width=\textwidth]{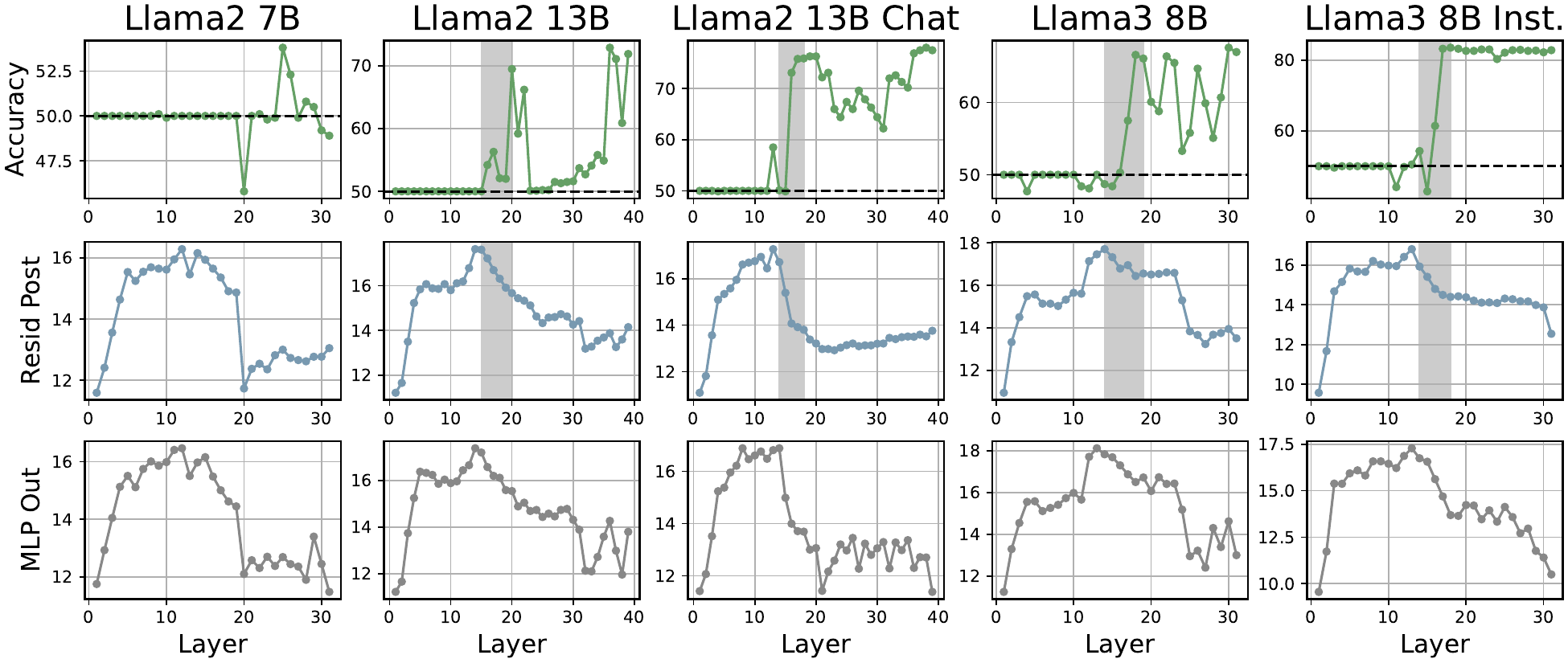}}
\vskip -0.1in
\caption[Accuracy and Intrinsic trends in COLD]{The figure shows accuracy along with ID trends for different variants of the LLama model on the \textbf{COLD} dataset}
\label{fig:peak-trend-COLD}
\end{center}
\vskip -0.35in
\end{figure*}

\begin{figure*}
\begin{center}
\centerline{\includegraphics[width=\textwidth]{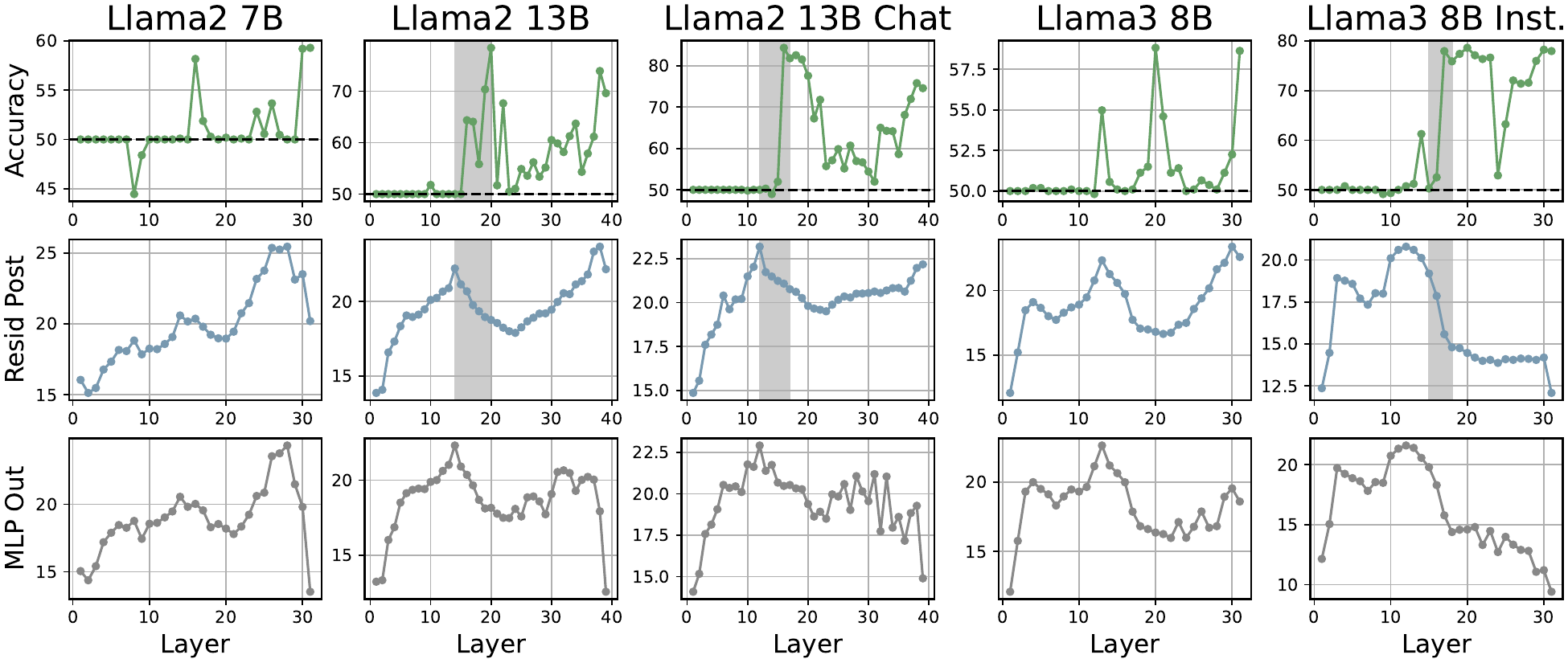}}
\vskip -0.1in
\caption[Accuracy and Intrinsic trends in Rotten Tomatoes]{The figure shows accuracy along with ID trends for different variants of the LLama model on the \textbf{Rotten Tomatoes} dataset}
\label{fig:peak-trend-RottenTomatoes}
\end{center}
\vskip -0.35in
\end{figure*}

\begin{figure*}
\begin{center}
\centerline{\includegraphics[width=\textwidth]{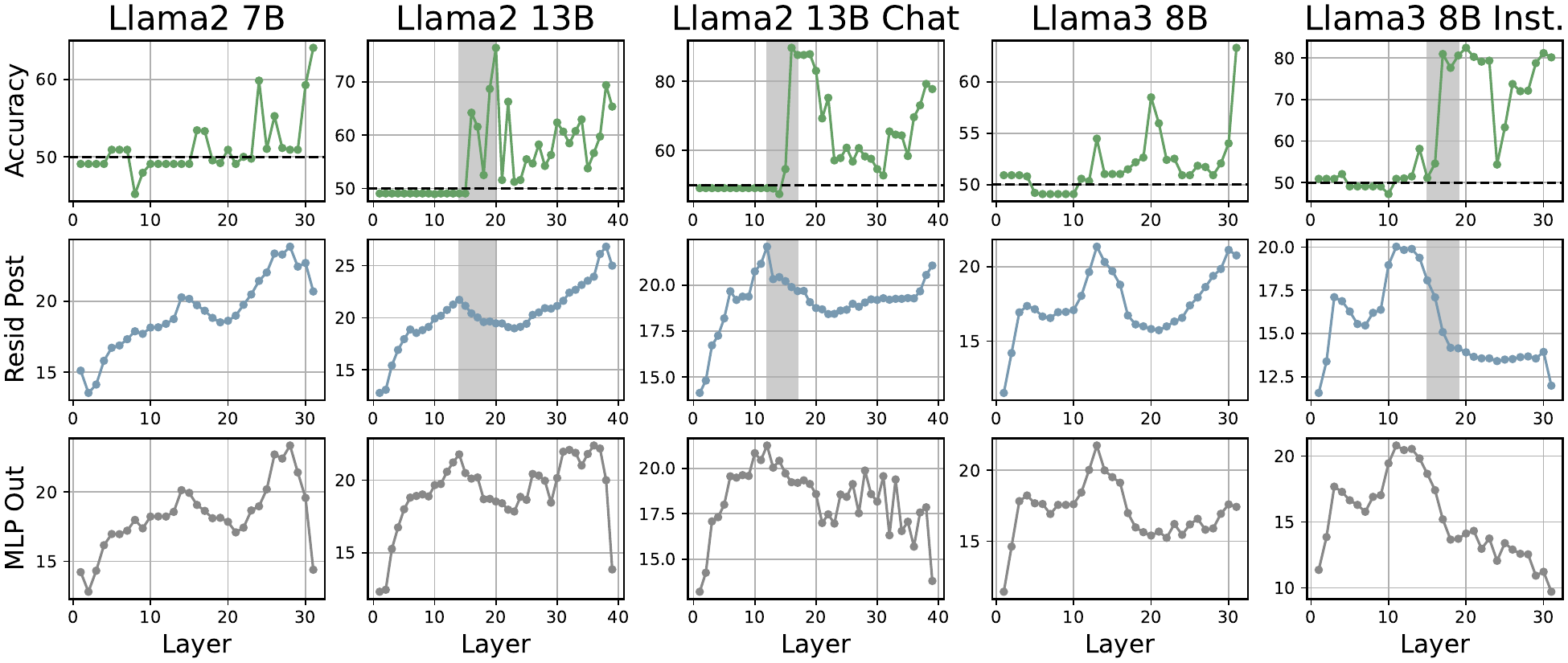}}
\vskip -0.1in
\caption[Accuracy and Intrinsic trends in SST2]{The figure shows accuracy along with ID trends for different variants of the LLama model on the \textbf{SST2} dataset}
\label{fig:peak-trend-SST2}
\end{center}
\vskip -0.35in
\end{figure*}

\begin{figure*}
\begin{center}
\centerline{\includegraphics[width=\textwidth]{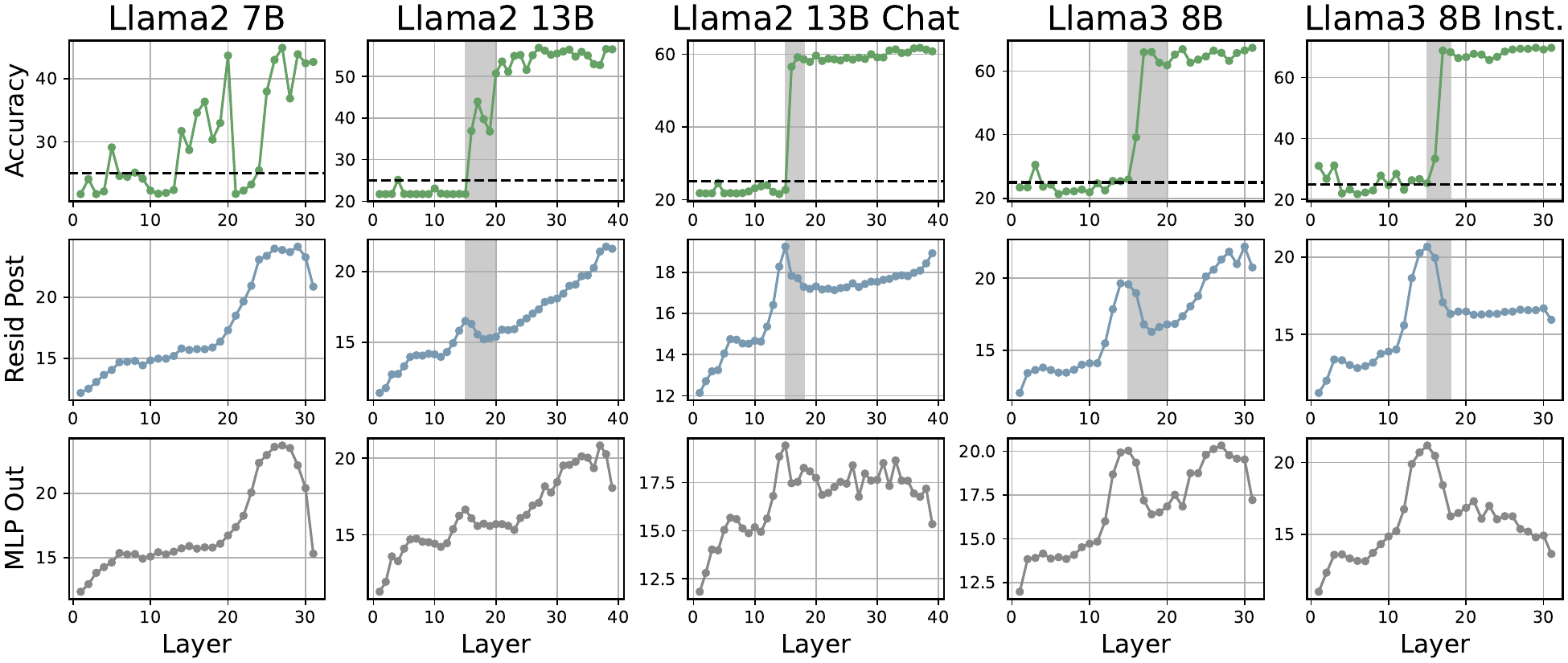}}
\vskip -0.1in
\caption[Accuracy and Intrinsic trends in MMLU Social Sciences]{The figure shows accuracy along with ID trends for different variants of the LLama model on the \textbf{MMLU Social Sciences} dataset}
\label{fig:peak-trend-MMLU-Social}
\end{center}
\vskip -0.35in
\end{figure*}

\begin{figure*}
\begin{center}
\centerline{\includegraphics[width=\textwidth]{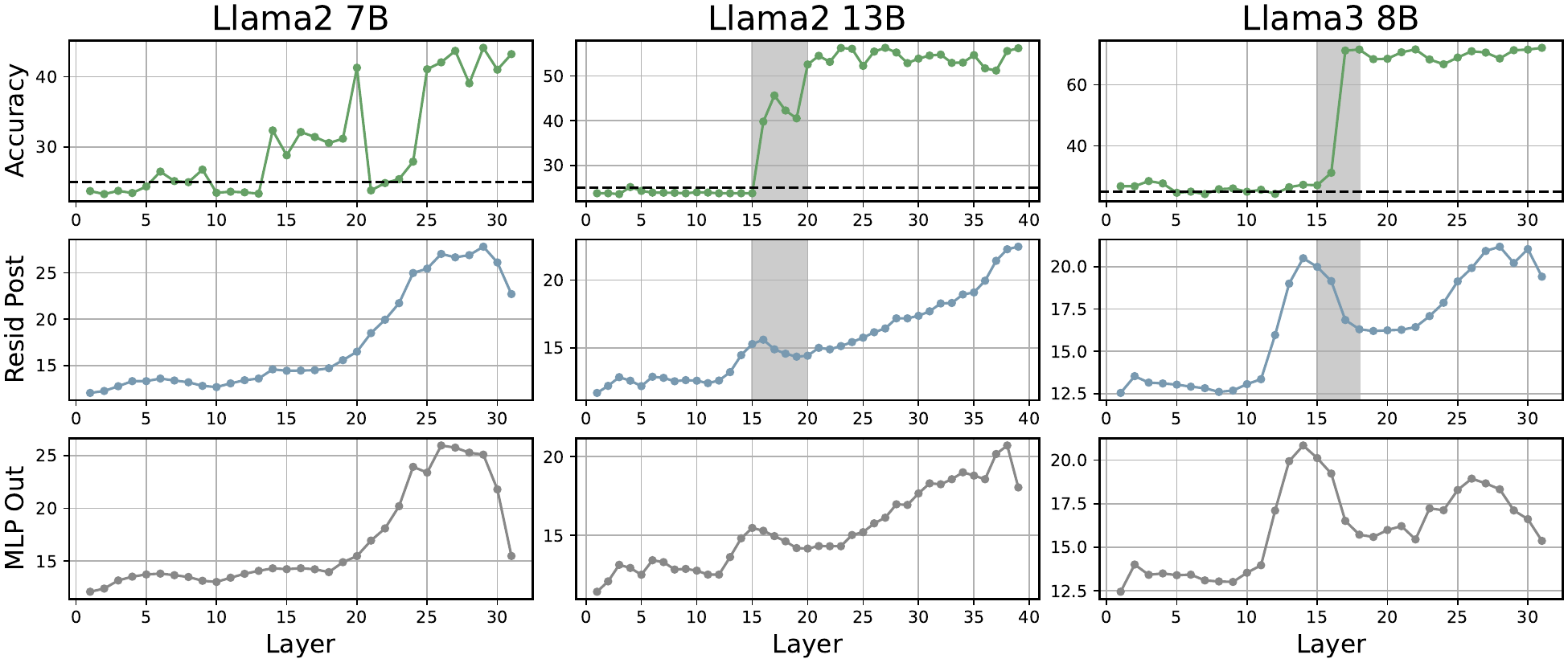}}
\vskip -0.1in
\caption[Accuracy and Intrinsic trends in MMLU Other]{The figure shows accuracy along with ID trends for different variants of the LLama model on the \textbf{MMLU Other} dataset}
\label{fig:peak-trend-MMLU-Other}
\end{center}
\vskip -0.35in
\end{figure*}

\begin{figure*}
\begin{center}
\centerline{\includegraphics[width=\textwidth]{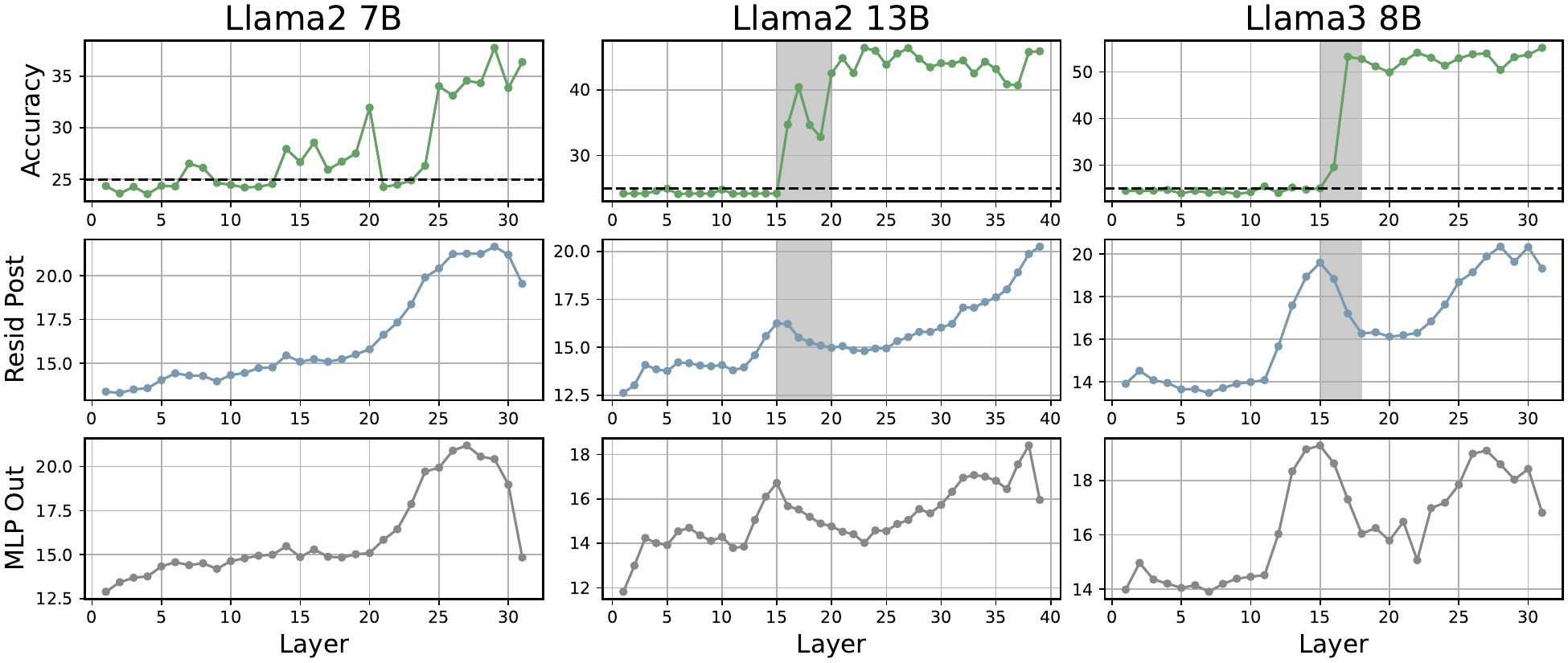}}
\vskip -0.1in
\caption[Accuracy and Intrinsic trends in MMLU Humanities]{The figure shows accuracy along with ID trends for different variants of the LLama model on the \textbf{MMLU Humanities} dataset}
\label{fig:peak-trend-MMLU-Humanities}
\end{center}
\vskip -0.35in
\end{figure*}

\begin{figure*}
\begin{center}
\centerline{\includegraphics[width=\textwidth]{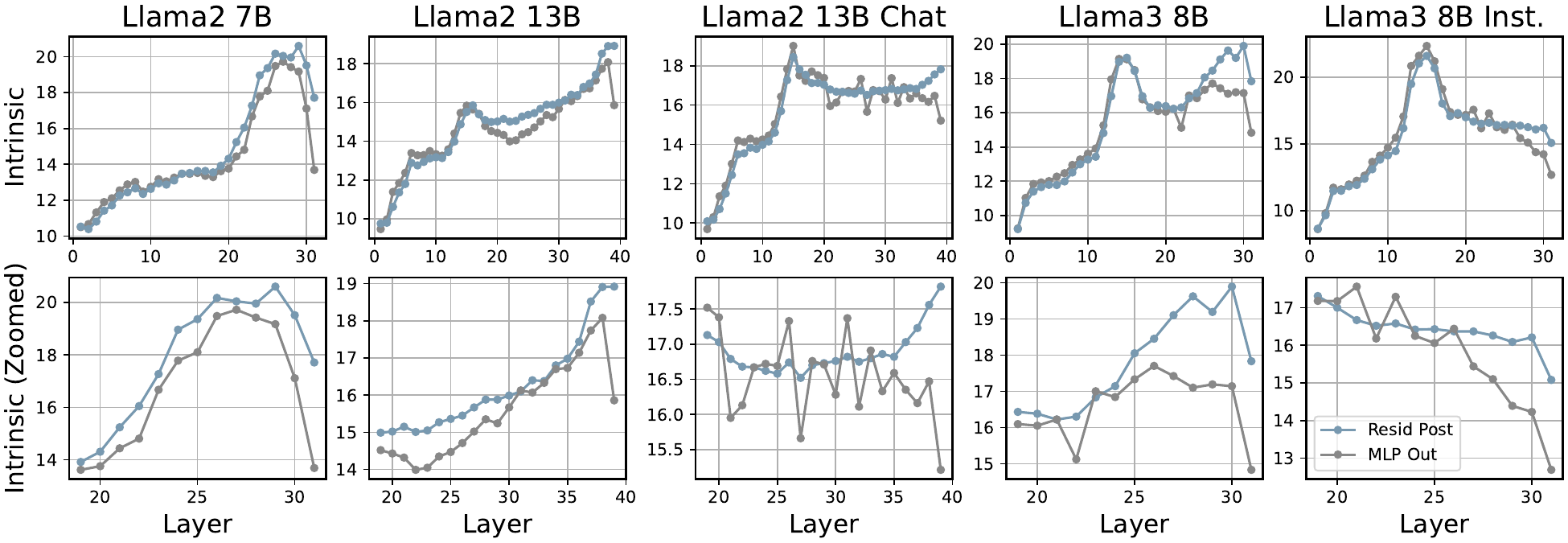}}
\vskip -0.1in
\caption[Residual Post Activation vs MLP Output Layer Intrinsic Estimates]{Comparision between ID across the model layer (MLP Out and Resid Post) on MMLU STEM Dataset}
\label{fig:resid_post_vs_mlp_out}
\end{center}
\vskip -0.35in
\end{figure*}

\begin{figure*}
\begin{center}
\centerline{\includegraphics[width=\textwidth]{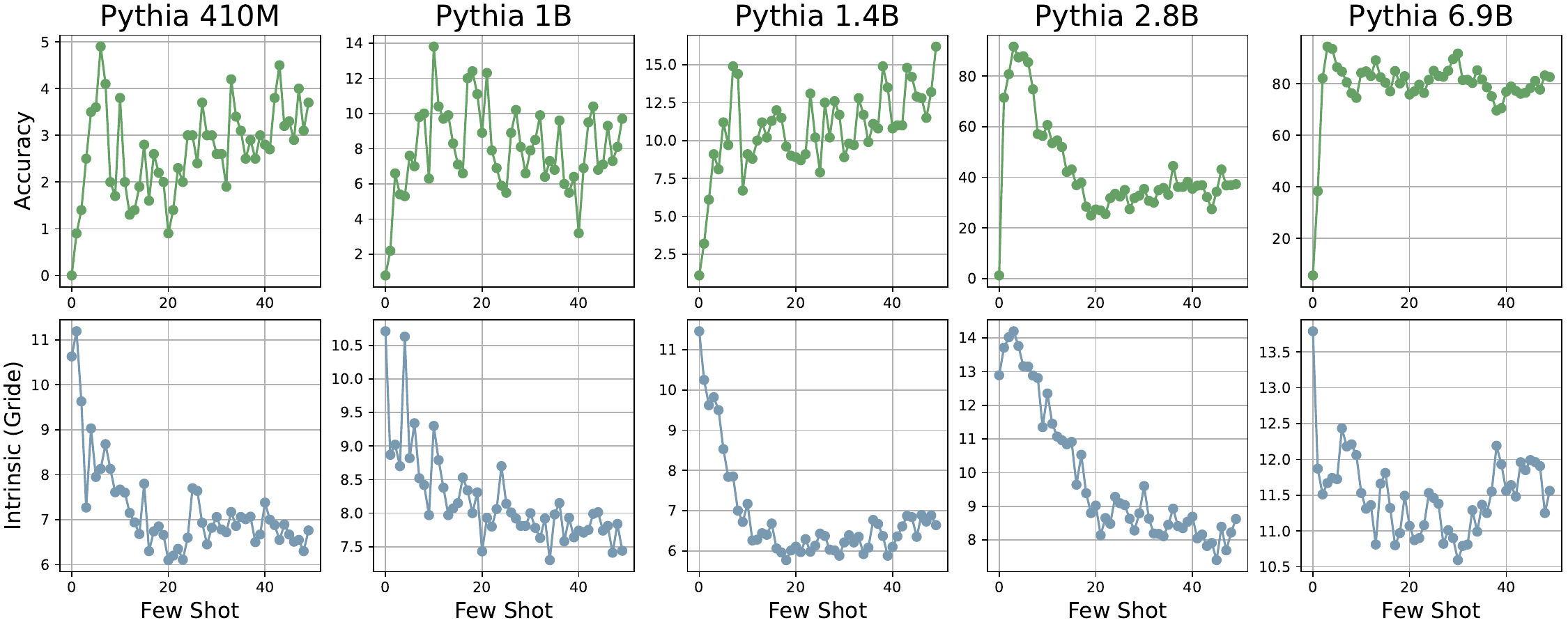}}
\vskip -0.1in
\caption[Few Shot 0 to 50 on Arithmetic Dataset, Pythia models]{Relationship between accuracy and intrinsic dimensionality across Pythia models of varying sizes on Arithmetic Dataset for few shots ranging from 0 to 50.}
\label{fig:few-shot-pythia}
\end{center}
\end{figure*}

\begin{figure*}
\begin{center}
\centerline{\includegraphics[width=\textwidth]{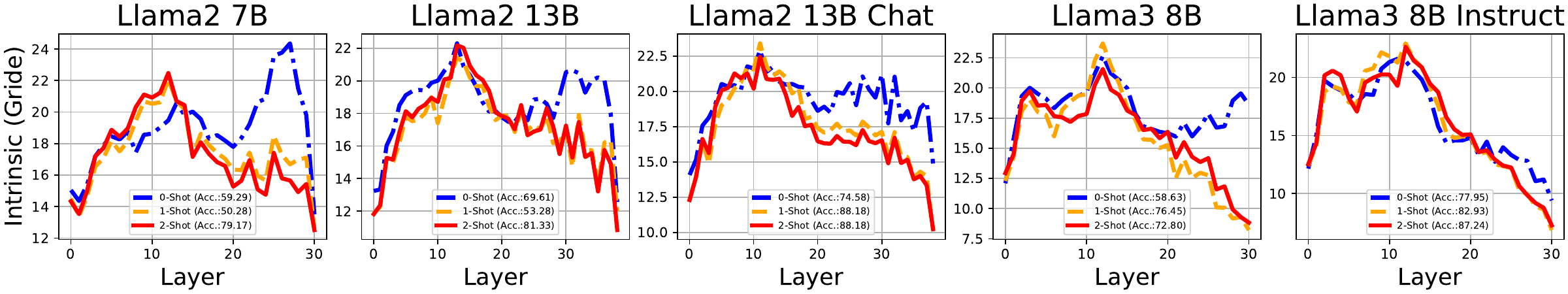}}
\caption[Few Shot trend in LLama models]{The figure shows ID of last token (MLP Out) throughout the stacked transformer blocks in the LLama family for the Rotten Tomatoes dataset. We observe that all these models show a similar characteristic curve of a hunchback-like shape, where the intrinsic dimensions first increase, reach a peak in the middle, and then further decrease. The legend includes the corresponding accuracies.
}
\label{app:fig:hunchback-rottentom}
\end{center}
\end{figure*}

\textbf{Comparison of trends in ID estimators:} 
In our experiments, we compute intrinsic dimensions using 3 widely used ID estimators (MLE, MLE-Modified, TwoNN) and include a recently introduced estimator (GRIDE). Overall, we found similar trends throughout multiple ID estimators. We stick to GRIDE estimates for more reliable results in the main paper. 

\input{./sections/discussion}

\begin{figure*}[ht]
\vskip 0.2in
\begin{center}
\scalebox{0.85}{
    \begin{tabular}{p{1.1\linewidth}}
      \toprule
      \texttt{Following are some multiple choice questions. You should directly answer the question by choosing the correct option.}\\
      \texttt{\textcolor{blue}{[ in-context examples (if few-shot/in-context learning experiment) ]}} \\
      \texttt{Question: \textcolor{teal}{A generalised statement pertaining to the task} -:  \textcolor{orange}{question/statement} 
      }\\
      \texttt{A. \textcolor{blue}{\textcolor{orange}{choice1}}} \\
      \texttt{B. \textcolor{blue}{\textcolor{orange}{choice2}}} \\
      \texttt{Answer:\textcolor{red}{ \underline{ A}}} \\
      \bottomrule
    \end{tabular}
    }
\caption[General Prompt Template]{Input prompt formats for the MCQA-based evaluation of autoregressive open-weight models (e.g., \texttt{llama(-2)}, \texttt{GPT-J}, etc.).
The \texttt{black text} is the templated input for all datasets. The \texttt{\textcolor{orange}{orange text}} is the input from the datasets, which contains either a review, a statement, or a question. 
The \texttt{\textcolor{teal}{teal text}} is a template comment describing the task, which changes according to the dataset
 The next-token prediction probabilities of the option IDs at the \textcolor{red}{\underline{\texttt{red text}}} are used as the observed prediction distribution. 
}
\label{app-fig:general-question-prompt}
\end{center}
\vskip -0.3in
\end{figure*}

\begin{figure*}[ht]
\vskip 0.2in
\begin{center}
\scalebox{0.85}{
    \begin{tabular}{p{1.1\linewidth}}
      \toprule
      \texttt{Following are some multiple choice questions. You should directly answer the question by choosing the correct option.}\\
      \texttt{\textcolor{blue}{[ in-context examples (if few-shot/in-context learning experiment) ]}} \\
      \texttt{Question: \textcolor{teal}{Which is best fitting topic for the given news report?} -: \textcolor{orange}{Brewers buyer expected to step out of the shadows Monday MILWAUKEE - Paul Attanasio says the story of his brother buying a baseball team is like a script straight out of Hollywood. He should know.} 
      }
      \\
      \texttt{A. \textcolor{blue}{\textcolor{orange}{World}}} \\
      \texttt{B. \textcolor{blue}{\textcolor{orange}{Sports}}} \\
      \texttt{C. \textcolor{blue}{\textcolor{orange}{Business}}} \\
      \texttt{D. \textcolor{blue}{\textcolor{orange}{Sci/Tech}}} \\
      \texttt{Answer:\textcolor{red}{ \underline{ C}}} \\
      \bottomrule
    \end{tabular}
    }
\caption[AG News dataset Prompt Template]{Input prompt formats for the MCQA-based evaluation of autoregressive open-weight models (e.g., \texttt{llama(-2)}, \texttt{GPT-J}, etc.).
The \texttt{black text} is the templated input for all datasets. The \texttt{\textcolor{orange}{orange text}} is the input from the \textbf{AG News dataset}.
The \texttt{\textcolor{teal}{teal text}} is a template comment describing the task.
 The next-token prediction probabilities of the option IDs at the \textcolor{red}{\underline{\texttt{red text}}} are used as the observed prediction distribution. }
\label{app-fig:agnews_question_prompt}
\end{center}
\vskip -0.3in
\end{figure*}

\begin{figure*}[ht]
\vskip 0.2in
\begin{center}
\scalebox{0.85}{
    \begin{tabular}{p{1.1\linewidth}}
      \toprule
      \texttt{Following are some multiple choice questions. You should directly answer the question by choosing the correct option.}\\
      \texttt{\textcolor{blue}{[ in-context examples (if few-shot/in-context learning experiment) ]}} \\
      \texttt{Question: \textcolor{orange}{Mars has an atmosphere that is almost entirely carbon dioxide. Why isn't there a strong greenhouse effect keeping the planet warm?} 
      }\\
      \texttt{A: \textcolor{blue}{\textcolor{orange}{the atmosphere on Mars is too thin to trap a significant amount of heat}}} \\
      \texttt{B: \textcolor{blue}{\textcolor{orange}{There actually is a strong greenhouse effect and Mars would be 35oC colder than it is now without it.}}} \\
      \texttt{C: \textcolor{blue}{\textcolor{orange}{Mars does not have enough internal heat to drive the greenhouse effect}}} \\
      \texttt{D: \textcolor{blue}{\textcolor{orange}{the greenhouse effect requires an ozone layer which Mars does not have}}} \\
      \texttt{Answer:\textcolor{red}{ \underline{ A}}} \\
      \bottomrule
    \end{tabular}
    }
\caption[MMLU dataset Prompt Template]{Input prompt formats for the MCQA-based evaluation of autoregressive open-weight models (e.g., \texttt{llama(-2)}, \texttt{GPT-J}, etc.).
The \texttt{black text} is the templated input for all datasets. The \texttt{\textcolor{orange}{orange text}} is the input from the \textbf{MMLU dataset}.
 The next-token prediction probabilities of the option IDs at the \textcolor{red}{\underline{\texttt{red text}}} are used as the observed prediction distribution. 
\label{app-fig:mmlu_question_prompt}
}
\end{center}
\vskip -0.3in
\end{figure*}


\begin{figure*}[ht]
\vskip 0.2in
\begin{center}
\scalebox{0.85}{
    \begin{tabular}{p{1.1\linewidth}}
      \toprule
      \texttt{Following are some multiple choice questions. You should directly answer the question by choosing the correct option.}\\
      \texttt{\textcolor{blue}{[ in-context examples (if few-shot/in-context learning experiment) ]}} \\
      \texttt{Question: \textcolor{teal}{Select the suitable option for the following statement} -: \textcolor{orange}{The cat was bitten the mouse.} 
      }\\
      \texttt{A: \textcolor{blue}{\textcolor{orange}{Unacceptable}}} \\
      \texttt{B: \textcolor{blue}{\textcolor{orange}{Acceptable}}} \\
      \texttt{Answer:\textcolor{red}{ \underline{ A}}} \\
      \bottomrule
    \end{tabular}
    }
\caption[CoLA dataset Prompt Template]{Input prompt formats for the MCQA-based evaluation of autoregressive open-weight models (e.g., \texttt{llama(-2)}, \texttt{GPT-J}, etc.).
The \texttt{black text} is the templated input for all datasets. The \texttt{\textcolor{orange}{orange text}} is the input from the \textbf{CoLA dataset}.
The \texttt{\textcolor{teal}{teal text}} is a template comment describing the task.
 The next-token prediction probabilities of the option IDs at the \textcolor{red}{\underline{\texttt{red text}}} are used as the observed prediction distribution. }
\label{app-fig:cola_question_prompt}
\end{center}
\vskip -0.3in
\end{figure*}


\begin{figure*}[ht]
\vskip 0.2in
\begin{center}
\scalebox{0.85}{
    \begin{tabular}{p{1.1\linewidth}}
      \toprule
      \texttt{Following are some multiple choice questions. You should directly answer the question by choosing the correct option.}\\
      \texttt{\textcolor{blue}{[ in-context examples (if few-shot/in-context learning experiment) ]}} \\
      \texttt{Question: \textcolor{teal}{Select the suitable option for the following statement} -: \textcolor{orange}{enchanted with low-life tragedy and liberally seasoned with emotional outbursts . . . what is sorely missing, however, is the edge of wild, lunatic invention that we associate with cage's best acting .} 
      }\\
      \texttt{A: \textcolor{blue}{\textcolor{orange}{Negative}}} \\
      \texttt{B: \textcolor{blue}{\textcolor{orange}{Positive}}} \\
      \texttt{Answer:\textcolor{red}{ \underline{ A}}} \\
      \bottomrule
    \end{tabular}
    }
\caption[Rotten Tomatoes dataset Prompt Template]{Input prompt formats for the MCQA-based evaluation of autoregressive open-weight models (e.g., \texttt{llama(-2)}, \texttt{GPT-J}, etc.).
The \texttt{black text} is the templated input for all datasets. The \texttt{\textcolor{orange}{orange text}} is the input from the \textbf{Rotten Tomatoes dataset}.
The \texttt{\textcolor{teal}{teal text}} is a template comment describing the task.
 The next-token prediction probabilities of the option IDs at the \textcolor{red}{\underline{\texttt{red text}}} are used as the observed prediction distribution. }
\label{app-fig:rottentomatoes_question_prompt}
\end{center}
\vskip -0.3in
\end{figure*}


\begin{figure*}[ht]
\vskip 0.2in
\begin{center}
\scalebox{0.85}{
    \begin{tabular}{p{1.1\linewidth}}
      \toprule
      \texttt{Following are some multiple choice questions. You should directly answer the question by choosing the correct option.}\\
      \texttt{\textcolor{blue}{[ in-context examples (if few-shot/in-context learning experiment) ]}} \\
      \texttt{Question: \textcolor{teal}{Select the suitable option for the following statement} -: \textcolor{orange}{this is human comedy at its most amusing, interesting and confirming .} 
      }\\
      \texttt{A: \textcolor{blue}{\textcolor{orange}{Negative}}} \\
      \texttt{B: \textcolor{blue}{\textcolor{orange}{Positive}}} \\
      \texttt{Answer:\textcolor{red}{ \underline{ B}}} \\
      \bottomrule
    \end{tabular}
    }
\caption[SST2 dataset Prompt Template]{Input prompt formats for the MCQA-based evaluation of autoregressive open-weight models (e.g., \texttt{llama(-2)}, \texttt{GPT-J}, etc.).
The \texttt{black text} is the templated input for all datasets. The \texttt{\textcolor{orange}{orange text}} is the input from the \textbf{SST-2 dataset}.
The \texttt{\textcolor{teal}{teal text}} is a template comment describing the task.
 The next-token prediction probabilities of the option IDs at the \textcolor{red}{\underline{\texttt{red text}}} are used as the observed prediction distribution. }
\label{app-fig:sst2_question_prompt}
\end{center}
\vskip -0.3in
\end{figure*}


\begin{figure*}[ht]
\vskip 0.2in
\begin{center}
\scalebox{0.85}{
    \begin{tabular}{p{1.1\linewidth}}
      \toprule
      \texttt{Following are some multiple choice questions. You should directly answer the question by choosing the correct option.}\\
      \texttt{\textcolor{blue}{[ in-context examples (if few-shot/in-context learning experiment) ]}} \\
      \texttt{Question: \textcolor{teal}{ Which of the following events (given as options A or B) is a more plausible} \textcolor{orange}{effect} \textcolor{teal}{of the event} -: \textcolor{orange}{'The woman betrayed her friend.'?} 
      }\\
      \texttt{A: \textcolor{blue}{\textcolor{orange}{Her friend sent her a greeting card.}}} \\
      \texttt{B: \textcolor{blue}{\textcolor{orange}{Her friend cut off contact with her.}}} \\
      \texttt{Answer:\textcolor{red}{ \underline{ B}}} \\
      \bottomrule
    \end{tabular}
    }
\caption[COPA dataset Prompt Template]{Input prompt formats for the MCQA-based evaluation of autoregressive open-weight models (e.g., \texttt{llama(-2)}, \texttt{GPT-J}, etc.).
The \texttt{black text} is the templated input for all datasets. The \texttt{\textcolor{orange}{orange text}} is the input from the \textbf{COPA dataset}.
The \texttt{\textcolor{teal}{teal text}} is a template comment describing the task.
 The next-token prediction probabilities of the option IDs at the \textcolor{red}{\underline{\texttt{red text}}} are used as the observed prediction distribution. }
\label{app-fig:copa_question_prompt}
\end{center}
\vskip -0.3in
\end{figure*}


\begin{figure*}[ht]
\vskip 0.2in
\begin{center}
\scalebox{0.85}{
    \begin{tabular}{p{1.1\linewidth}}
      \toprule
      \texttt{Following are some multiple choice questions \textcolor{teal}{about the activity 'going grocery shopping'}. You should directly answer the question by choosing the correct option.}\\
      \texttt{\textcolor{blue}{[ in-context examples (if few-shot/in-context learning experiment) ]}} \\
      \texttt{Question: \textcolor{teal}{ Which of the following events (given as options A or B) is a more plausible} \textcolor{orange}{cause} \textcolor{teal}{of the event} \textcolor{orange}{'drive to the nearby store.'?} 
      }\\
      \texttt{A: \textcolor{blue}{\textcolor{orange}{ make a list.}}} \\
      \texttt{B: \textcolor{blue}{\textcolor{orange}{get into car.}}} \\
      \texttt{Answer:\textcolor{red}{ \underline{ B}}} \\
      \bottomrule
    \end{tabular}
    }
\caption[COLD dataset Prompt Template]{Input prompt formats for the MCQA-based evaluation of autoregressive open-weight models (e.g., \texttt{llama(-2)}, \texttt{GPT-J}, etc.).
The \texttt{black text} is the templated input for all datasets. The \texttt{\textcolor{orange}{orange text}} is the input from the \textbf{COLD dataset}.
The \texttt{\textcolor{teal}{teal text}} is a template comment describing the task.
 The next-token prediction probabilities of the option IDs at the \textcolor{red}{\underline{\texttt{red text}}} are used as the observed prediction distribution. }
\label{app-fig:cold_question_prompt}
\end{center}
\vskip -0.3in
\end{figure*}

%% file: sections/discussion.tex


%
%

\subsection{Future Directions}
The analysis we perform raises some interesting directions for future analysis. 1) Throughout all the observations, we found a space with a similar intrinsic dimension created by these models across different layers. We observe a peak arising in all these models in the middle layers. Though our findings suggest the peak pointing towards the space where the model starts to be decisive, a more detailed analysis of this space for different skills would be an interesting future venue, exploring if there lie multiple manifolds for multiple skills/tasks. 
2) Provided the extrinsic to intrinsic dimension being large (found across multiple experiments), the autoregressive training objective highlights the knowledge compression capabilities of these models, making justifications for model compression by weight pruning/low-rank adaptations/finetuning/knowledge distillation. This also opens up the requirement for a more detailed analysis of manifolds evolving during low-rank finetuning objectives. 
3) Though we find that the low-dimension manifolds learned by different models lie in similar ranges, little is explored about their structural similarity in low-dimensional manifolds learned by different models.
In the future, it would be interesting to compare these subspaces in a more rigorous fashion, including comparison with subspace matching methods like Grassmann distance \cite{ye2016schubertvarietiesdistancessubspaces}.
4) Study of intrinsic dimensions changing via in-context learning examples provided in the prompt. Though the initial findings suggest the reduction in the manifold space as more in-context examples are provided, a more detailed study would be required to exploit ID estimates for choosing better in-context examples. 
5) Comparison of representational space of different modalities in vision-language models. In this work, we only focused on language modality, both in terms of language models and datasets. Some of the findings have suggested that networks learning similar representations for multiple modalities \cite{pmlr-v235-huh24a-platonic-representation-hypothesis, valeriani2024geometry}; these findings could be reinforced by observing the manifolds learned for different modalities by the same models. 6) Though our work highlights the ID estimates showing a strong relation with model generalization, exploiting them to develop a concrete unsupervised algorithm for model comparison/task complexity comparison and generated text comparison remains open for future avenues. 7) Some of the preliminary findings also suggest ID estimates of human text be different from LLM-generated text \cite{tulchinskii2024intrinsic}. In this work, we could only explore open-weight LLMs on English datasets, leaving the extended comparison in different languages for the future.